\documentclass[10pt,twocolumn,letterpaper]{article}

\usepackage{cvpr}
\usepackage{times}
\usepackage{epsfig}
\usepackage{graphicx}
\usepackage{amsmath}
\usepackage{amssymb}

\usepackage{epsfig}
\usepackage{graphicx}
\usepackage{amsmath}
\usepackage{amssymb}
\usepackage{tabularx}
\usepackage{multirow}
\usepackage{amssymb}
\usepackage{amsmath}
\usepackage{pifont}
\usepackage{subfigure}

\usepackage[super]{nth}
\newcommand{\cmark}{\ding{51}}%
\newcommand{\xmark}{\ding{55}}%
\usepackage{color}
\usepackage{balance}
\usepackage{booktabs}
\usepackage{overpic}
\newcommand\mypara[1]{\vspace{1mm}\noindent\textbf{#1}}

\usepackage[breaklinks=true,bookmarks=false]{hyperref}

\cvprfinalcopy 

\def\cvprPaperID{2094} 

\begin{document}

\title{Flow2Stereo: Effective Self-Supervised Learning of \\ Optical Flow and Stereo Matching}

\author{Pengpeng Liu$^\dag$\thanks{Work mainly done during an internship at Huya AI.}  \qquad\
Irwin King$^\dag$  \qquad\
Michael Lyu$^\dag$ \qquad\
Jia Xu$^\S$ \\
$^\dag$  The Chinese University of Hong Kong \qquad\
$^\S$ Huya AI \\
}

\maketitle

\begin{abstract}
In this paper, we propose a unified method to jointly learn optical flow and stereo matching. Our first intuition is stereo matching can be modeled as a special case of optical flow, and we can leverage  3D geometry behind stereoscopic videos to guide the learning of these two forms of correspondences. We then enroll this knowledge into the  state-of-the-art self-supervised learning framework, and train one single network to estimate both flow and stereo. Second, we unveil the bottlenecks in prior self-supervised learning approaches, and propose  to create a new set of challenging proxy tasks to boost performance. These two insights  yield a single model that achieves the highest accuracy among all existing unsupervised flow and stereo methods  on KITTI 2012 and  2015  benchmarks. More remarkably, our self-supervised method even outperforms several state-of-the-art fully supervised methods, including PWC-Net and FlowNet2 on KITTI 2012.
\end{abstract}

\section{Introduction}
Estimating optical flow and stereo matching are two fundamental computer vision tasks with a wide range of applications~\cite{geiger2012we,menze2015object}.
 Despite impressive progress in the past decades, accurate flow and stereo estimation remain a long-standing  challenge. Traditional stereo matching estimation approaches often employ different pipelines  compared with prior   flow estimation methods~\cite{horn1981determining,brox2011large,revaud2015epicflow,kanade1991stereo,sun2003stereo,scharstein2002taxonomy,hirschmuller2007evaluation,hirschmuller2008stereo,geiger2010efficient}. These methods merely share common modules, and they are  computationally expensive.

Recent CNN-based methods  directly estimate optical flow~\cite{dosovitskiy2015flownet,ilg2017flownet,ranjan2017optical,sun2018pwc,hui18liteflownet} or stereo matching \cite{kendall2017end,chang2018pyramid} from two raw images, achieving high accuracy with real-time speed. However, these fully supervised methods require a large amount of labeled data to obtain  state-of-the-art performance. Moreover, CNNs for flow estimation  are  drastically different from those for stereo estimation in terms of network architecture and training data~\cite{dosovitskiy2015flownet,Mayer_2016_CVPR}.

\begin{figure}[t]
\centering
\includegraphics[width=\linewidth]{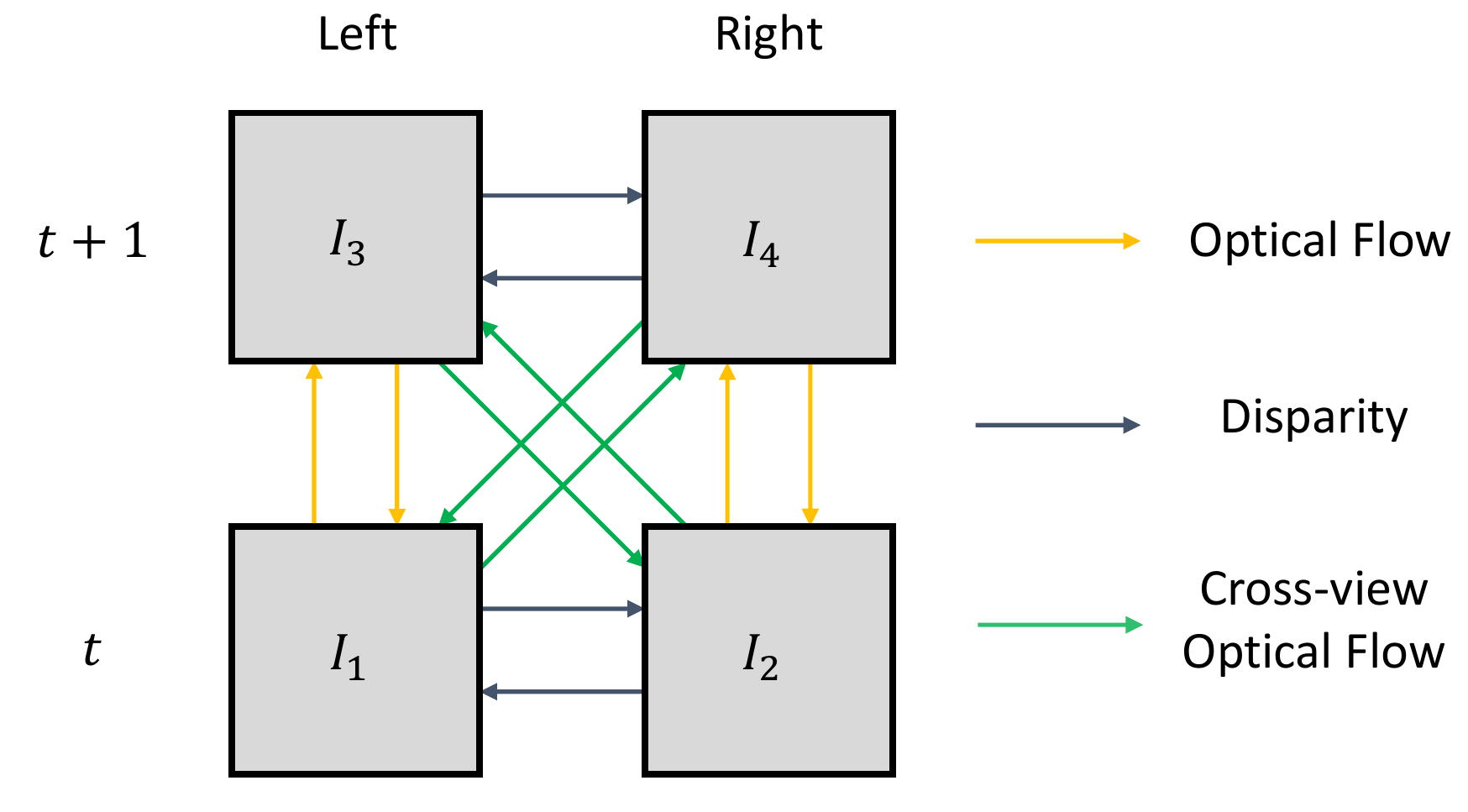} 
\caption{Illustration of $12$   cross-view  correspondnce maps among $4$ stereoscopic frames. We leverage all these geometric consistency constraints, and train one single network to estimate both flow and stereo.}
\label{teaser}
\end{figure}

Is it possible to train one single network to estimate both flow and stereo  using only one set of data,  even unlabeled?
In this paper, we show conventional self-supervised  methods can learn to estimate these two forms  of dense correspondences with one single model, when  fully utilizing  stereoscopic videos with inherent geometric constraints.

\begin{figure}[t]
\centering
\includegraphics[width=0.96\linewidth]{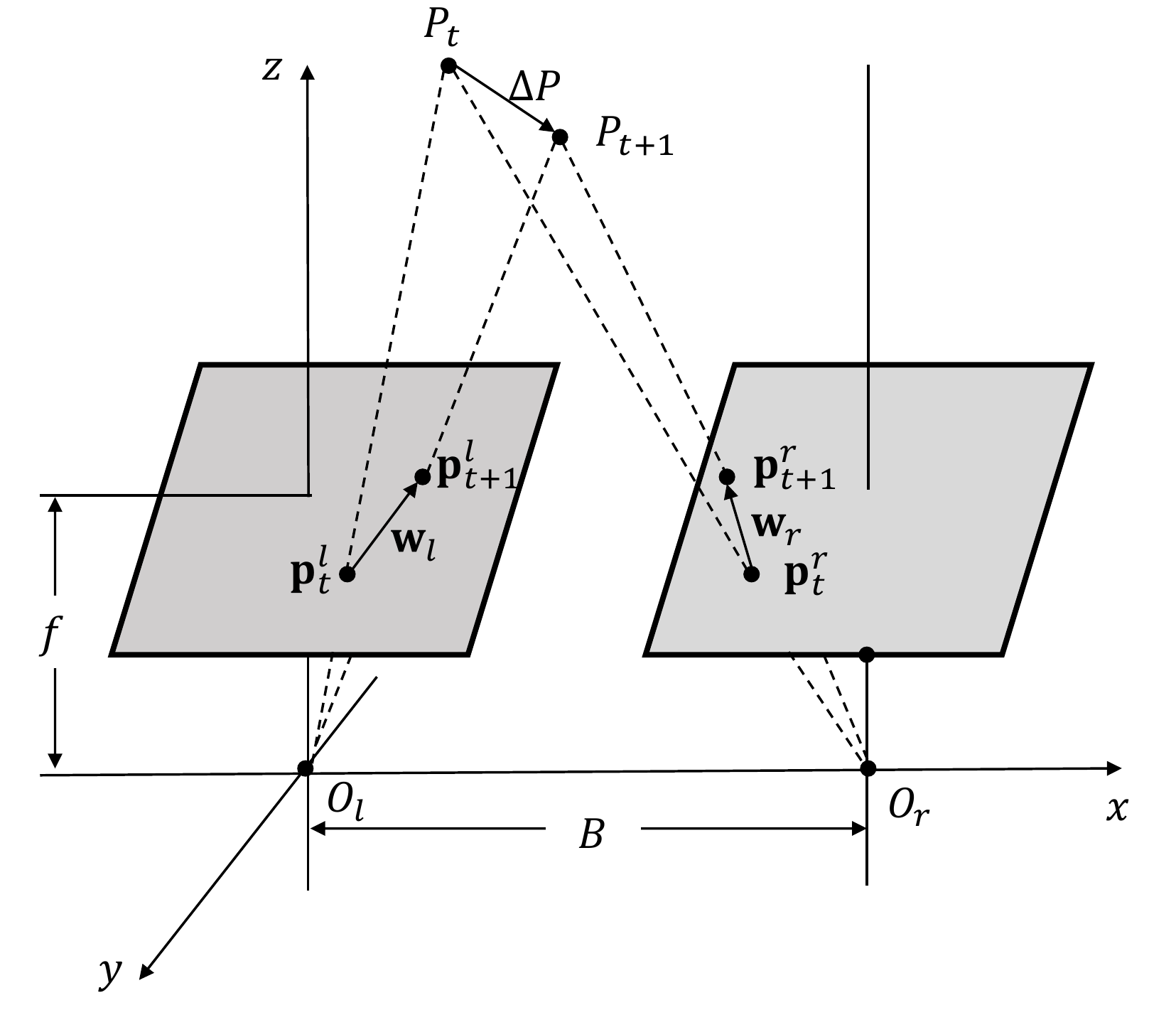} 
\caption{3D geometric constraints between optical flow ($\textbf{w}_l$ and $\textbf{w}_r$) and stereo disparity from time $t$ to $t+1$ in the 3D projection view.}
\label{Epipolar}
\end{figure}

Fig.~\ref{Epipolar} shows the geometric relationship between stereo disparity and optical flow. We consider stereo matching as a special case of optical flow, and compute all $12$ cross-view  correspondence maps between images captured at different time and different view (as shown in Fig.~\ref{teaser}). This enables us to  train one single network   with a set of     photometric and geometric consistency constraints. Moreover, after digging into conventional two-stage self-supervised learning framework~\cite{Liu:2019:DDFlow,Liu:2019:SelFlow}, we find creating challenging proxy tasks is the key for performance improvement. Based on this observation, we propose to employ additional challenging conditions to further boost the performance.

These two insights yield a method outperforming all existing unsupervised flow learning methods by a large margin, with $Fl-noc = 4.02\%$ on KITTI 2012 and $Fl-all = 11.10\%$ on KITTI 2015. Remarkably, our self-supervised method even outperforms several state-of-the-art fully supervised  methods, including PWC-Net~\cite{sun2018pwc}, FlowNet2~\cite{ilg2017flownet},    and MFF~\cite{ren2018fusion} on KITTI 2012. More importantly, when we directly estimate  stereo matching with our optical flow model, it also achieves state-of-the-art unsupervised stereo matching performance. This further demonstrates  the strong generalization capability of our approach.

\section{Related Work}
Optical flow  and stereo matching have been widely studied in the past decades~\cite{horn1981determining,brox2011large,revaud2015epicflow,sun2010secrets,XRK2017,Zabih:1994:NLT,Zhang2009CrossBasedLS,Mei_2013_CVPR, klaus2006segment,hirschmuller2008stereo}. Here, we briefly review recent deep learning based methods.

\mypara{Supervised Flow Methods.}  FlowNet~\cite{dosovitskiy2015flownet}  is the  first end-to-end optical learning method, which takes two raw images as input and output a dense flow map. The followup FlowNet 2.0~\cite{ilg2017flownet} stacks several basic FlowNet models and refines the flow iteratively, which significantly improves accuracy. SpyNet~\cite{ranjan2017optical}, PWC-Net~\cite{sun2018pwc} and LiteFlowNet~\cite{hui18liteflownet} propose to warp CNN features instead of image at different scales and introduce cost volume construction,   achieving state-of-the-art performance with a compact model size. However, these supervised methods rely on pre-training on synthetic datasets due to lacking of real-world ground truth optical flow. The very recent SelFlow~\cite{Liu:2019:SelFlow} employs self-supervised pre-training with real-world unlabeled data before fine-tuning, reducing the reliance of synthetic datasets. In this paper, we propose an unsupervised method, and  achieve comparable performance with supervised learning methods without using any labeled data.

\mypara{Unsupervised \& Self-Supervised Flow Methods.} Labeling optical flow for real-world images is a challenging task, and recent studies turn to formulate optical flow estimation as an unsupervised learning problem based on the brightness constancy and spatial smoothness assumption~\cite{jason2016back,ren2017unsupervised}. \cite{Meister:2018:UUL,wang2018occlusion,Janai2018ECCV} propose to detect occlusion and exclude occluded pixels when computing photometric loss. Despite promising progress, they still lack the ability to learn optical flow of occluded pixels.

Our work is most similar to DDFlow~\cite{Liu:2019:DDFlow} and SelFlow~\cite{Liu:2019:SelFlow}, which  employ a two-stage self-supervision strategy to cope with optical flow of occluded pixels.  In this paper, we extend the scope to utilize geometric constraints in stereoscopic videos
and   jointly learn optical flow and stereo disparity. This turns out to be very effective, as our method significantly improves the quality of flow prediction in the first stage.

Recent works also propose to jointly learn flow and depth from monocular videos~\cite{zhou2017unsupervised,zou2018df,yin2018geonet,ranjan2019competitive,liu2019unsupervised} or jointly learn flow and disparity from stereoscopic videos~\cite{lai2019bridging,wang2019unos}. Unlike these methods, we make full use of  geometric constraints between optical flow and stereo matching   in a self-supervised learning manner, and achieve much better performance.

\mypara{Unsupervised \& Self-Supervised Stereo Methods.}
Our method is also related to a large body of unsupervised stereo learning methods, including image synthesis and warping with depth estimation~\cite{garg2016unsupervised}, left-right consistency~\cite{Godard_2017_CVPR,Zhou_2017_ICCV,Guo_2018_ECCV}, employing  additional semantic information~\etal~\cite{yang2018segstereo}, cooperative learning~\cite{li2018occlusion},  self-adaptive fine-tuning~\etal\cite{tonioni2017unsupervised,zhong2018open,tonioni2019real}. Different from all these methods that design a specific network for stereo estimation, we train one single unified network to estimate both flow and stereo.

\begin{figure*}[t]
\centering
\includegraphics[width=0.95\textwidth]{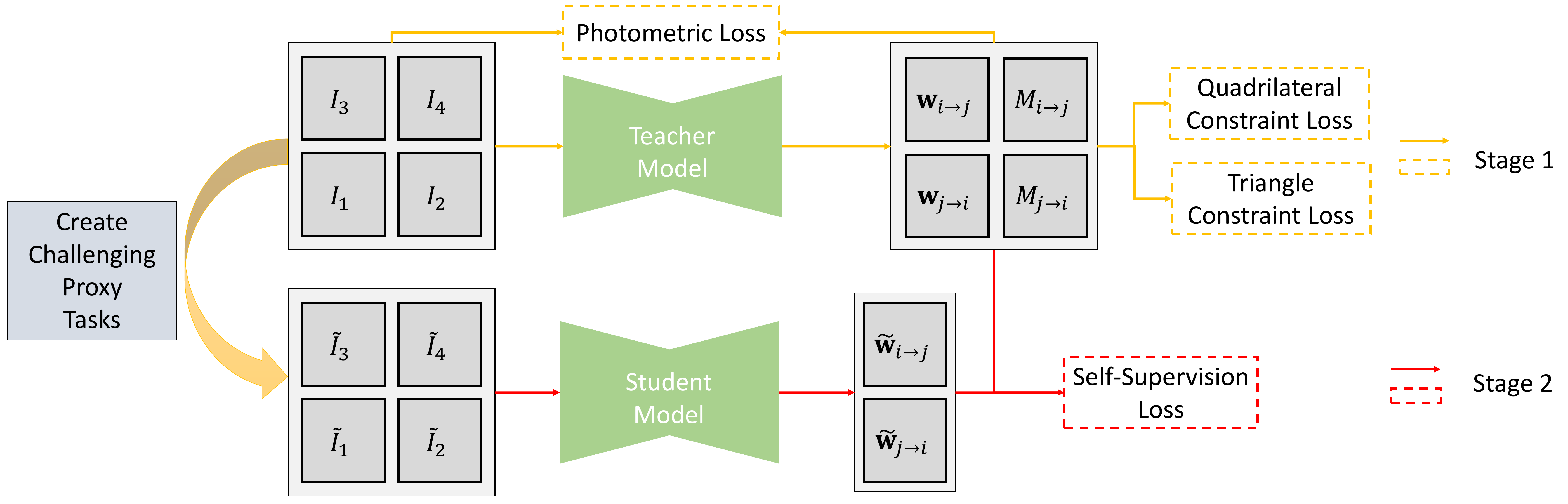}\\
\caption{Our  self-supervised learning framework contains two stages: In stage 1, we add geometric constraints between optical flow and stereo disparity to improve the quality of confident predictions; In stage 2, we create challenging proxy tasks to guide the student model for effective self-supervised learning.}
\label{Structure}
\end{figure*}

\section{Geometric Relationship of Flow and Stereo}
In this section, we review the geometric relationship between optical flow and stereo disparity from both the 3D projection view~\cite{hangeometric} and the motion view.

\subsection{Geometric Relationship in 3D  Projection}
Fig.~\ref{Epipolar} illustrates the geometric relationship between stereo disparity and optical flow from a 3D projection view. $O_l$ and $O_r$ are rectified left and right camera centers, $B$ is the baseline distance between two camera centers.

Suppose $P(X, Y, Z)$ is a 3D point at time $t$, and it moves to $P+\Delta P$ at time $t+1$, resulting in the displacement as $\Delta P = (\Delta X, \Delta Y, \Delta Z)$. Denote $f$ as the focal length, $p=(x, y)$ as the projection point of $P$ on the image plane, then $(x, y) = \frac {f} {s} \frac {(X, Y)} {Z}$, where $s$ is the scale factor that converts the world space   to the pixel space, \ie, \ how many meters per pixel. For simplicity, let $f' = \frac {f} {s}$, we have $ (x, y) = {f'} \frac {(X, Y)} {Z}$.
Take the time derivative, we obtain
\begin{equation}
\frac {(\Delta x, \Delta y)} {\Delta t} = f'\frac {1} {Z} \frac {(\Delta X, \Delta Y)} {\Delta t} - f'\frac {(X, Y)} {Z^2} \frac {\Delta Z} {\Delta t}
\label{eq: 02}
\end{equation}
Let $\textbf{w} = (u, v)$ be the optical flow vector ($u$ denotes motion in the $x$ direction and $v$ denotes motion in the $y$ direction) and the time step is one unit (from $t$ to $t+1$), then  Eq.~(\ref{eq: 02}) becomes,
\begin{equation}
(u, v) = f'\frac {(\Delta X, \Delta Y)} {Z} -  f' \frac {\Delta Z}{Z^2} {(X, Y)}
\label{eq: 03}
\end{equation}
For calibrated stereo cameras, we let $P$ in   the coordinate system of $O_l$. Then $P_l = P = (X, Y, Z)$ in the coordinate system of $O_l$ and $P_r = (X-B, Y, Z)$ in the coordinate system of $O_r$. With Eq.~(\ref{eq: 03}), we  obtain,
\begin{equation}
\begin{cases}
&(u_l, v_l) = f'\frac {(\Delta X, \Delta Y)} {Z} -  f' \frac {\Delta Z}{Z^2}  {(X, Y)} \\
&(u_r, v_r) = f'\frac {(\Delta X, \Delta Y)} {Z} -  f' \frac {\Delta Z}{Z^2}  {(X-B, Y)}
\end{cases},
\label{eq: 04}
\end{equation}
This can be further simplified as,
\begin{equation}
\begin{cases}
&u_r - u_l = f'B \frac {\Delta Z}{Z^2}  \\
&v_r - v_l = 0
\end{cases},
\label{eq: 05}
\end{equation}
Suppose $d$ is the stereo disparity ($d\ge0$), according to the depth $Z$ and the distance between two camera centers $B$, we have
$ d = f' \frac{B} {Z}$. Take the time derivative, we have
\begin{equation}
\frac {\Delta d} {\Delta t} = -f'\frac B {Z^2} \frac {\Delta Z}{\Delta t}
\label{eq: 07}
\end{equation}
Similarly, we set time step to be one unit, then
\begin{equation}
d_{t+1} -d_t = -f'B \frac {\Delta Z}{Z^2}
\label{eq: 08}
\end{equation}
With Eq.~(\ref{eq: 05}) and (\ref{eq: 08}), we  finally obtain,
\begin{equation}
\begin{cases}
&u_r - u_l = (-d_{t+1}) - (-d_t)  \\
&v_r - v_l = 0
\end{cases}.
\label{eq: 09}
\end{equation}
 Eq.  (\ref{eq: 09}) demonstrates the 3D geometric relationship between optical flow and stereo disparity, \ie, the difference between optical flow from left and right view is equal to the difference between disparity from time $t$ to $t+1$. Note that Eq.~(\ref{eq: 09}) also works when  cameras move, including rotating and translating from $t$ to $t+1$. Eq.~(\ref{eq: 09}) assumes the focal length is fixed, which is common for stereo cameras.

 \begin{figure*}[t]
 \centering
   \begin{tabular}{@{}c@{}c@{}c@{}c@{}c@{}}
 \begin{overpic} [width=0.195\textwidth]{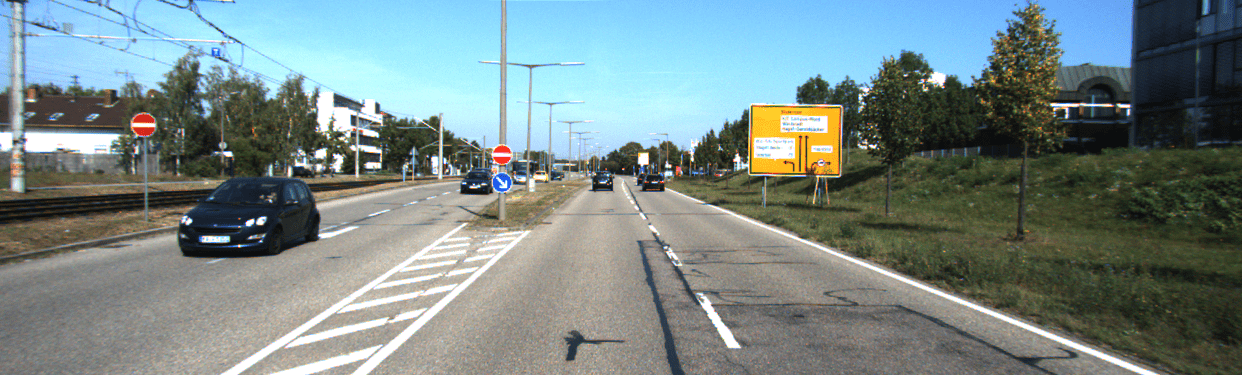} \put(4, 20){\color{red} \textbf{$I_t$}} \end{overpic} &
 \includegraphics[width=0.195\textwidth]{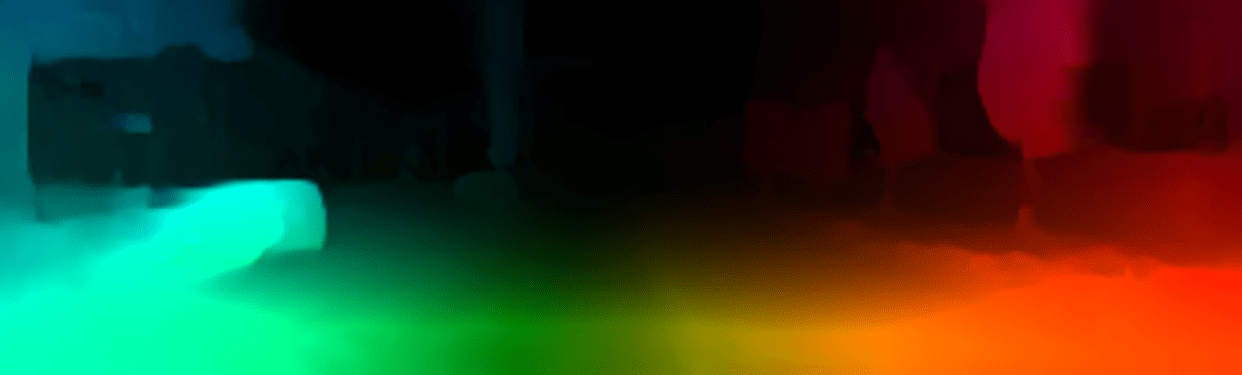} &
 \includegraphics[width=0.195\textwidth]{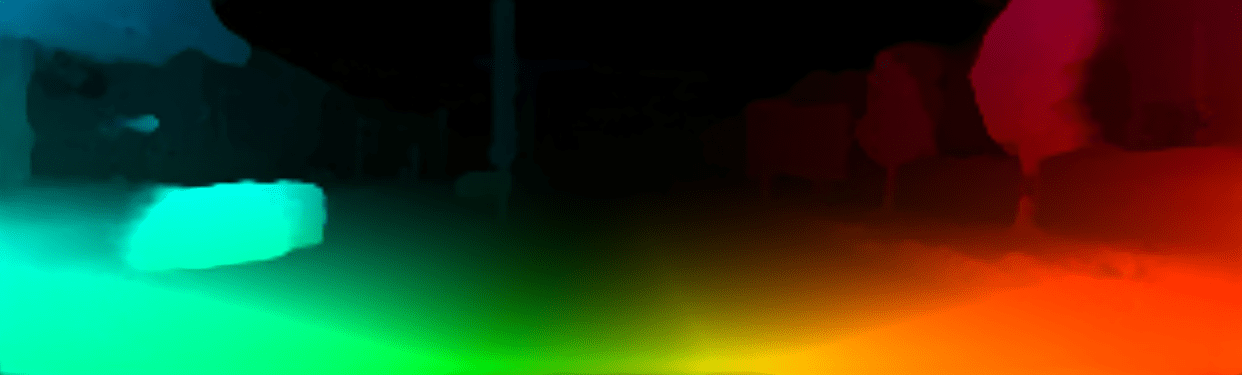} &
 \includegraphics[width=0.195\textwidth]{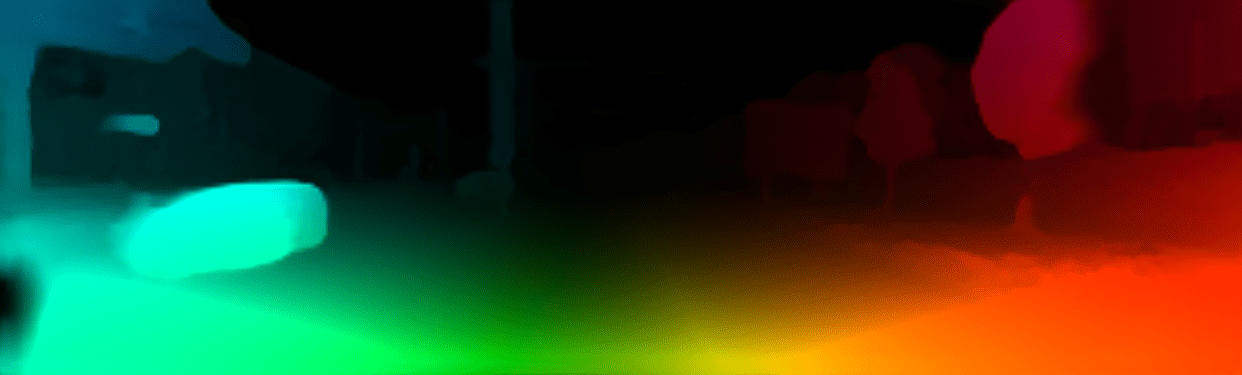} &
 \includegraphics[width=0.195\textwidth]{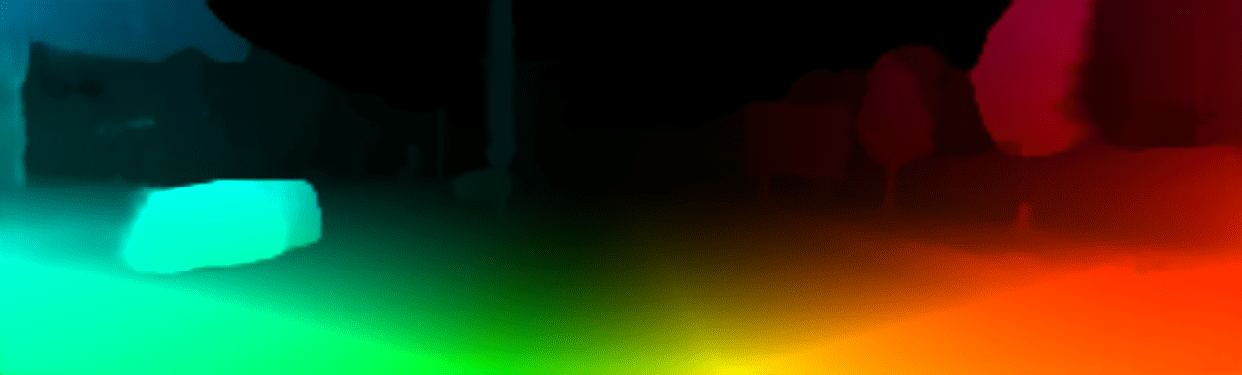} \\
 \begin{overpic} [width=0.195\textwidth]{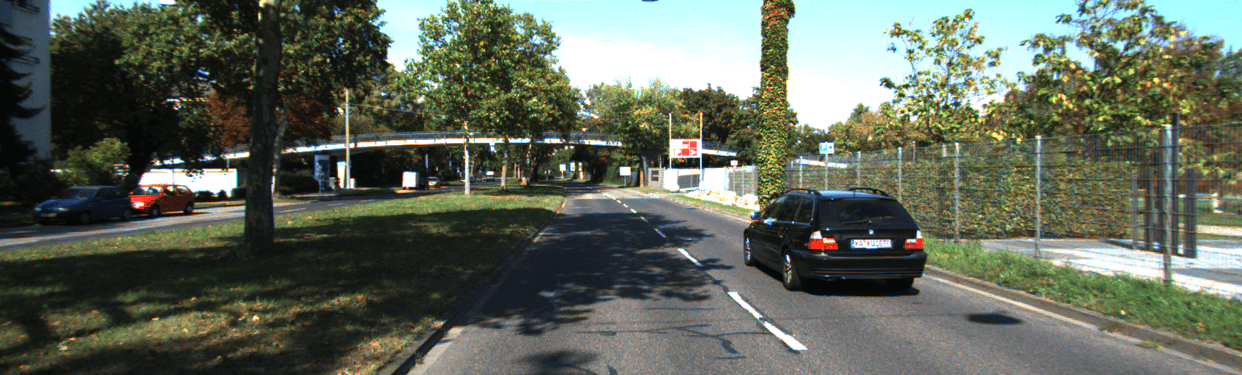} \put(4, 20){\color{red} \textbf{$I_{t+1}$}} \end{overpic}&
 \begin{overpic} [width=0.195\textwidth]{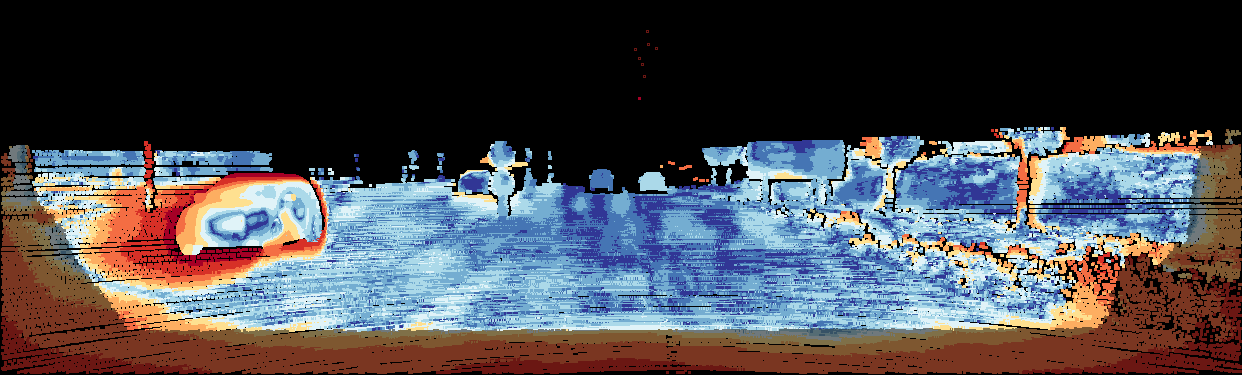} \put(4, 20){\color{white} \textbf{Fl: 31.39\%}} \end{overpic}&
 \begin{overpic} [width=0.195\textwidth]{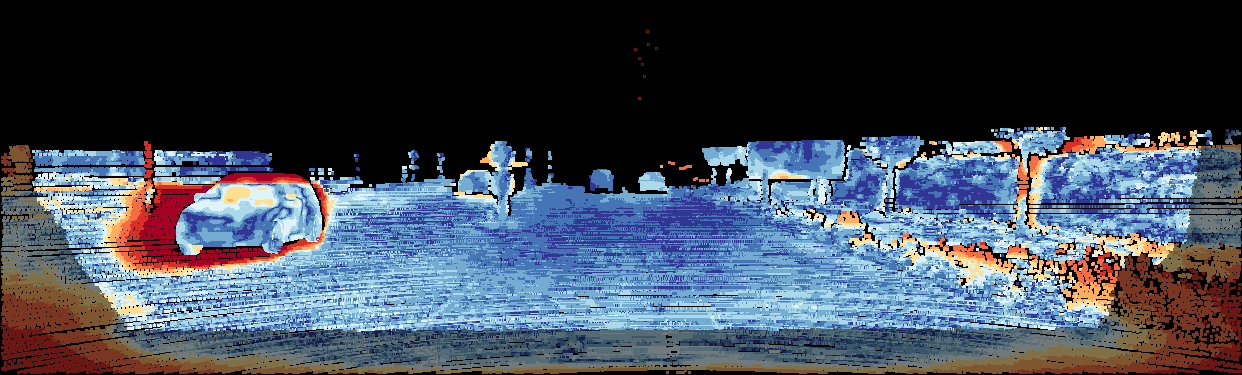} \put(4, 20){\color{white} \textbf{Fl: 14.38\%}} \end{overpic}&
 \begin{overpic} [width=0.195\textwidth]{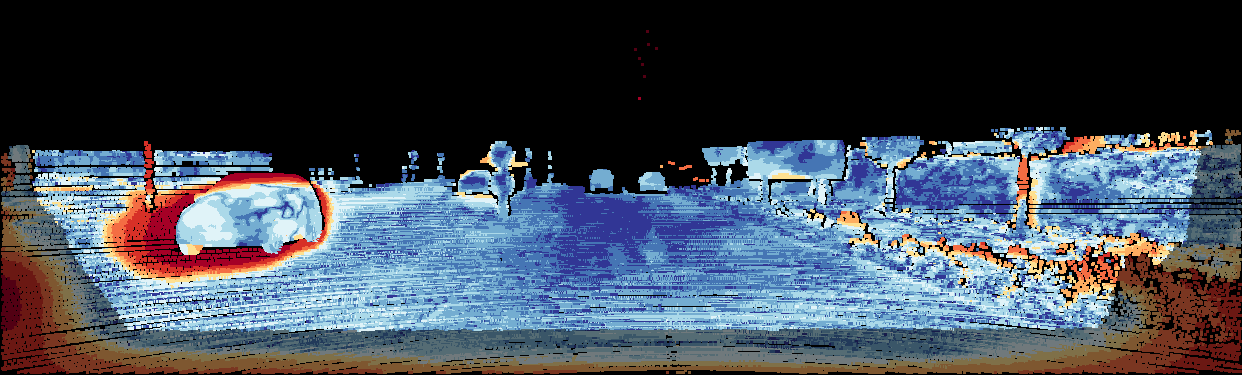}\put(4, 20){\color{white} \textbf{Fl: 15.19\%}} \end{overpic}&
 \begin{overpic} [width=0.195\textwidth]{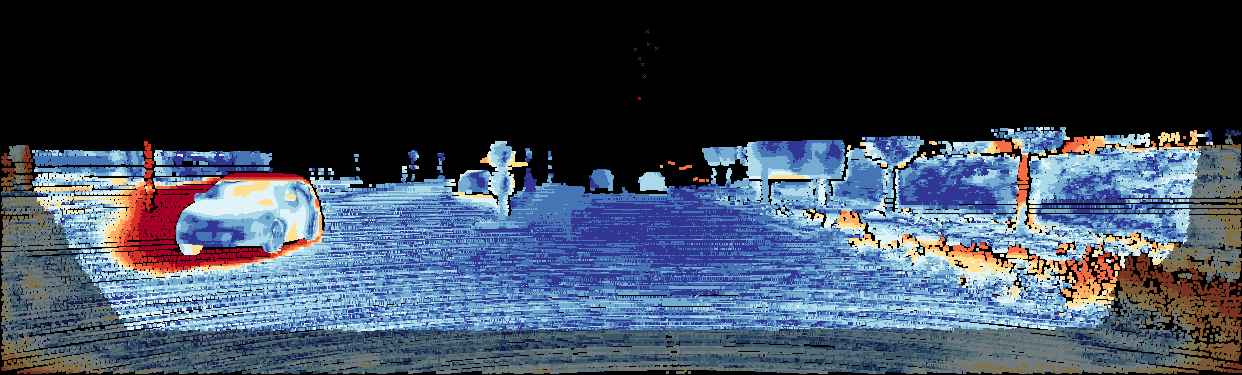}\put(4, 20){\color{white} \textbf{Fl: 8.31\%}} \end{overpic}\\

 \begin{overpic} [width=0.195\textwidth]{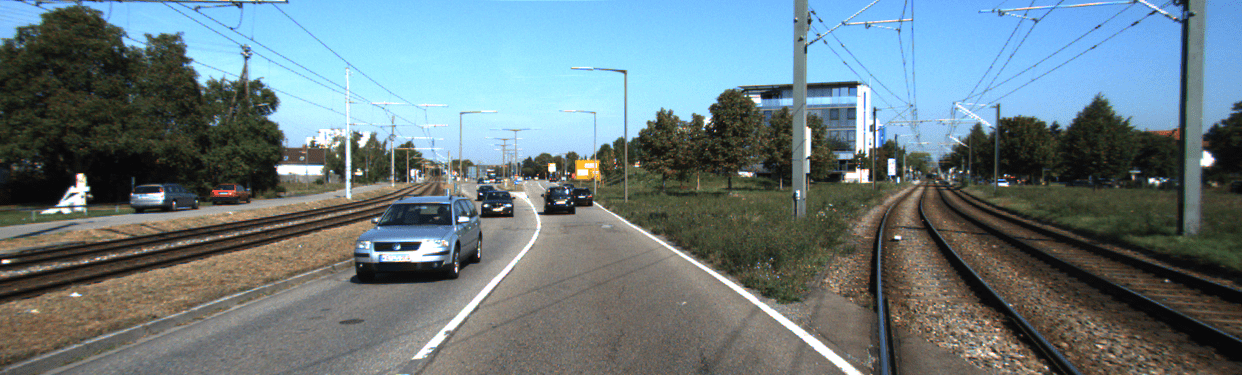} \put(4, 20){\color{red} \textbf{$I_t$}} \end{overpic} &
 \includegraphics[width=0.195\textwidth]{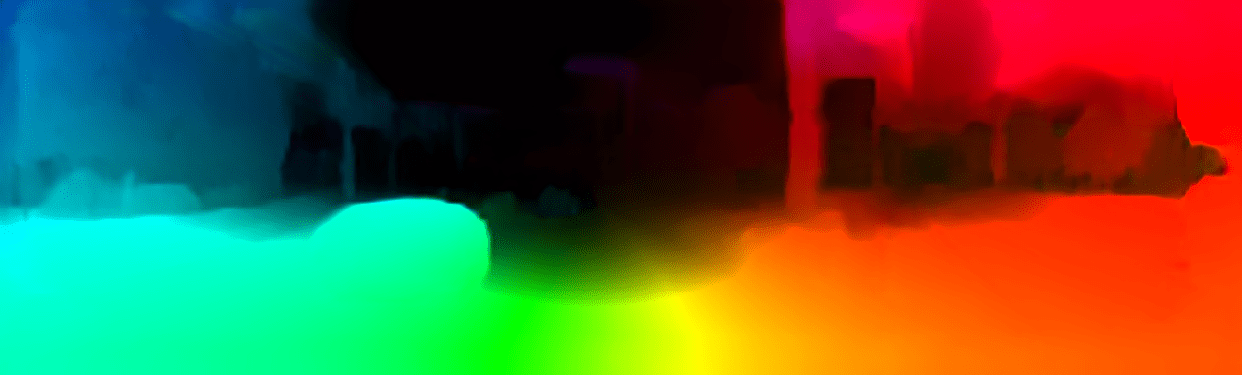} &
 \includegraphics[width=0.195\textwidth]{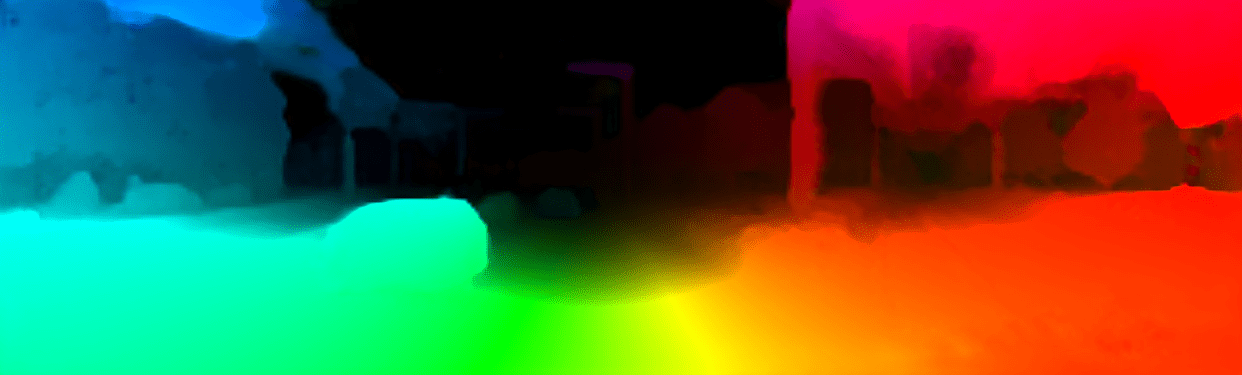} &
 \includegraphics[width=0.195\textwidth]{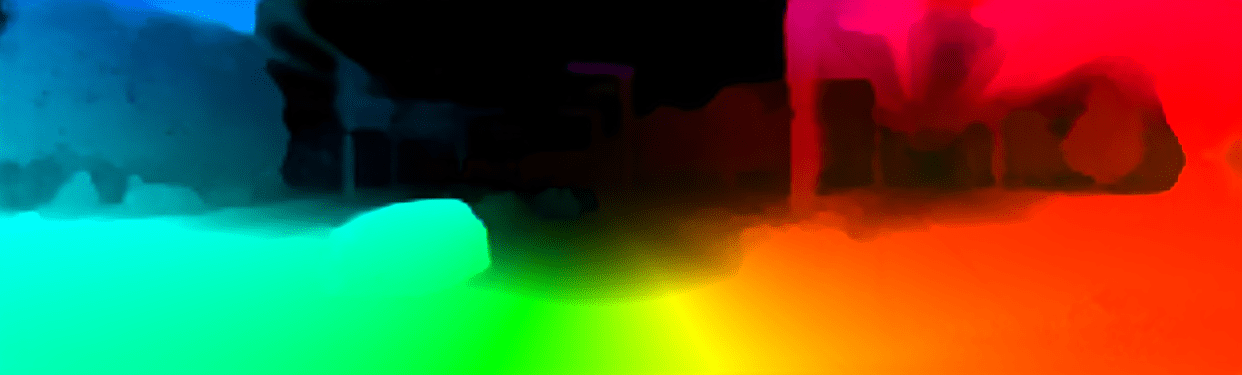} &
 \includegraphics[width=0.195\textwidth]{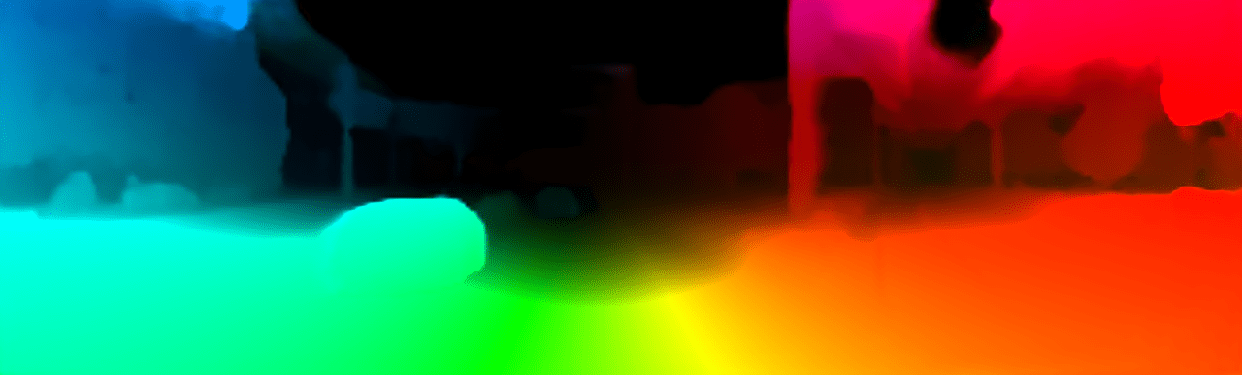} \\
 \begin{overpic} [width=0.195\textwidth]{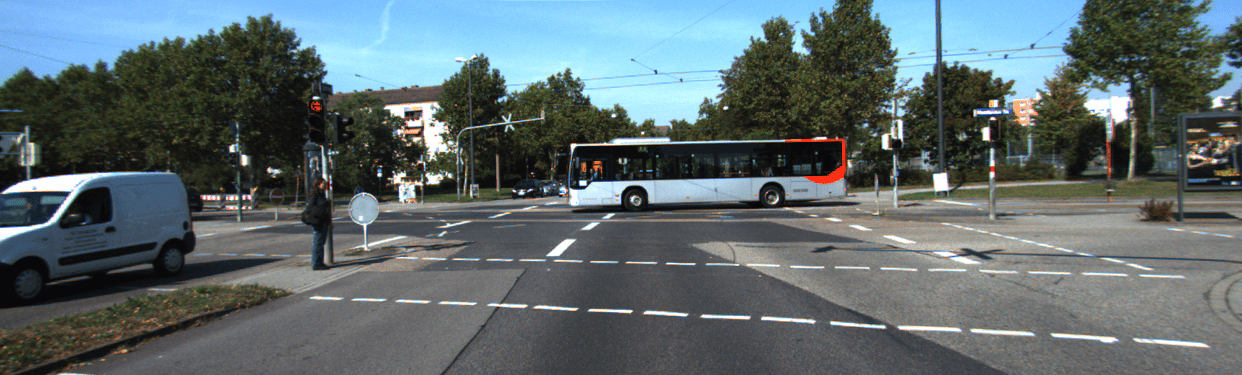} \put(4, 20){\color{red} \textbf{$I_{t+1}$}} \end{overpic} &
 \begin{overpic} [width=0.195\textwidth]{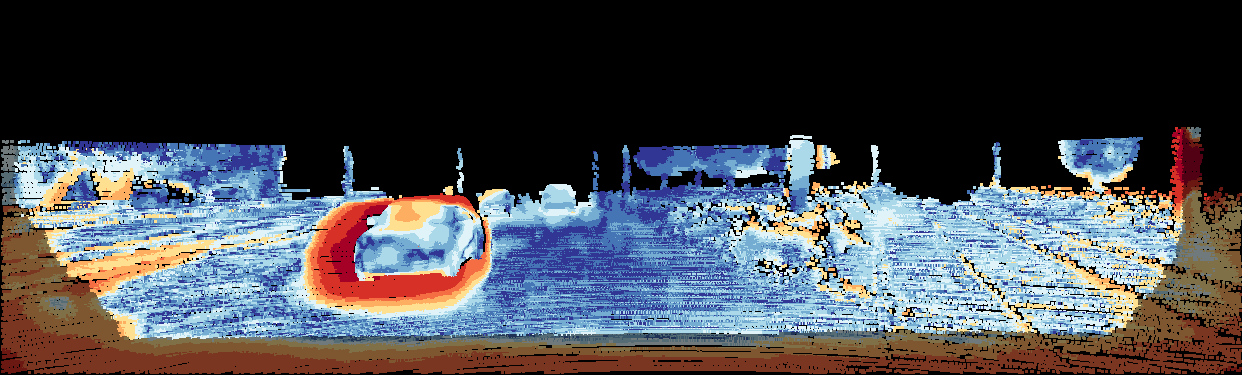} \put(4, 20){\color{white} \textbf{Fl: 27.14\%}} \end{overpic} &
 \begin{overpic} [width=0.195\textwidth]{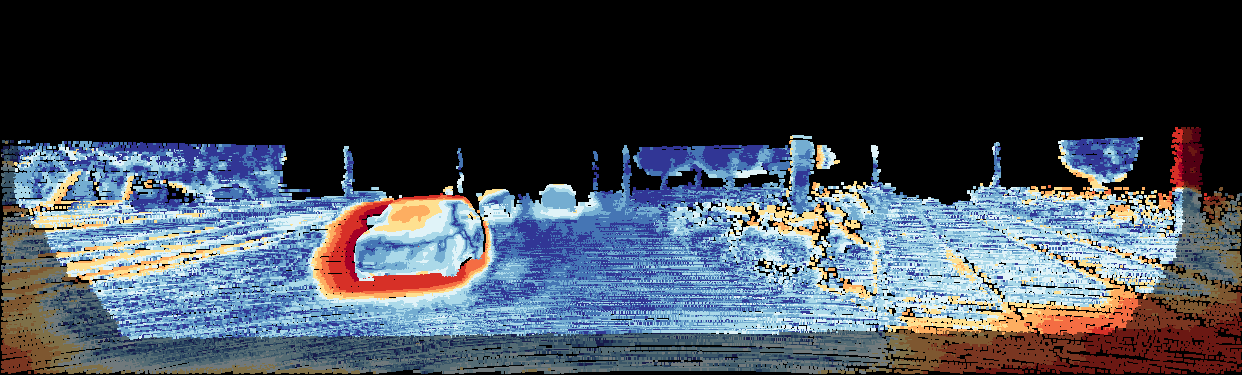} \put(4, 20){\color{white} \textbf{Fl: 15.07\%}} \end{overpic}  &
 \begin{overpic} [width=0.195\textwidth]{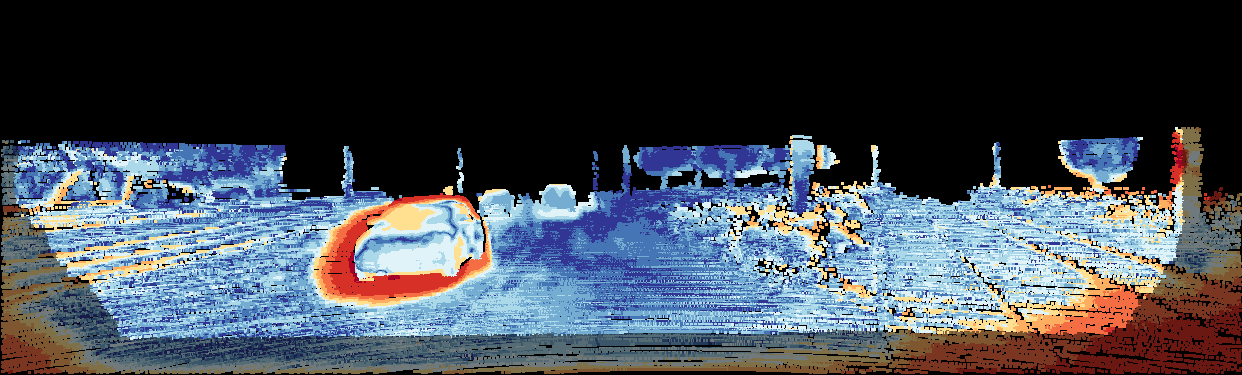} \put(4, 20){\color{white} \textbf{Fl: 16.40\%}} \end{overpic} &
 \begin{overpic} [width=0.195\textwidth]{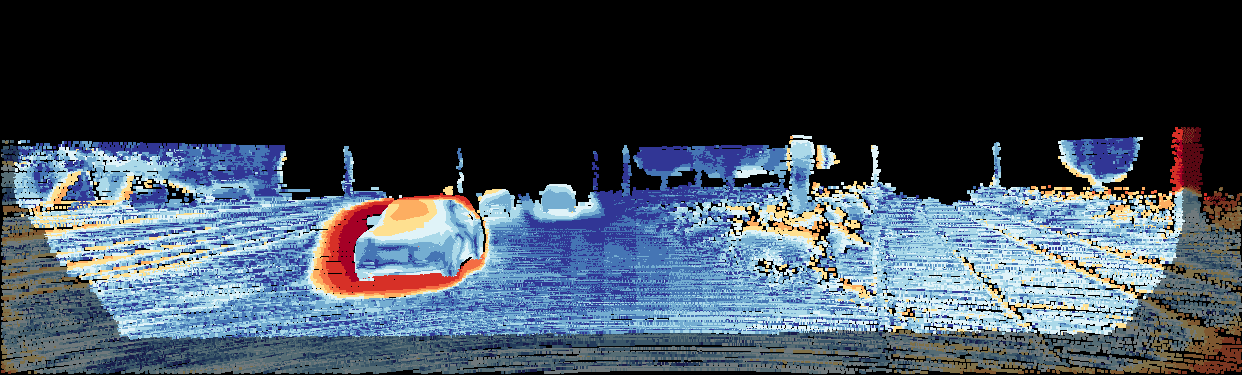}\put(4, 20){\color{white} \textbf{Fl: 9.59\%}} \end{overpic}\\
 (a) Input Images & (b) MFO-Flow~\cite{Janai2018ECCV}  & (c) DDFlow~\cite{Liu:2019:DDFlow}  & (d) SelFlow~\cite{Liu:2019:SelFlow} & (e) Ours\\
   \end{tabular}
 \caption{Qualitative evaluation on KITTI 2015 optical flow benchmark. For each case, the top row is optical flow and the bottom row is error map. Our model achieves much better results both quantitatively and qualitatively (\eg, shaded boundary regions). Lower Fl is better.}
 \label{BenchmarkFlow}
 \end{figure*}

Next, we review the geometric relationship between flow and stereo in the motion view.

\subsection{Geometric Relationship in Motion }
Optical flow estimation and stereo matching can be viewed as one single problem: correspondence  matching.
Optical flow  describes the pixel motion between two adjacent frames recorded at different time, while stereo disparity represents the pixel displacement  between two stereo images recorded at the same time. According to stereo geometry, the correspondence pixel shall lie on the epipolar line between stereo images. However, optical flow does not have such a constraint, this is because both camera and object can move at different times.

To this end,  we  consider stereo matching as a special case of optical flow. That is, the displacement between stereo images can be seen as a one dimensional  movement. For rectified stereo image pairs, the epipolar line is horizontal and stereo matching becomes finding the correspondence pixel along the horizontal direction $x$.

In the following, we consider stereo disparity as a form of motion between stereo image pairs.
For simplicity, let $I_1, I_3$ denote the left-view images at time $t$ and $t+1$, $I_2, I_4$ denote the right-view images at time $t$ and $t+1$ respectively. Then we let $\textbf{w}_{1 \to 3}$ denote the optical flow from $I_1$ to $I_3$,  $\textbf{w}_{1 \to 2}$ denote the stereo disparity from $I_1$ to $I_2$. For stereo disparity, we only keep the horizontal direction of optical flow. For optical flow and disparity of other directions, we denote them in the same way.

Apart from optical flow in the left and right view, disparity at time $t$ and $t+1$, we also compute the cross-view optical flow between images captured at different time and different view, such as $\textbf{w}_{1 \to 4}$ (green row in Fig.~\ref{teaser}). In this case, we compute the correspondence between every two images, resulting in $12$ optical flow maps as shown in Fig.~\ref{teaser}. We employ the same model to compute optical flow between every two images.

Suppose $\textbf{p}_t^l$ is a pixel in $I_1$, $\textbf{p}_t^r, \textbf{p}_{t+1}^{l}, \textbf{p}_{t+1}^{r}$ are its correspondence pixels in $I_2$, $I_3$ and $I_4$ respectively, then we have,
\begin{equation}
\begin{cases}
&\textbf{p}_t^r = \textbf{p}_t^l + \textbf{w}_{1 \to 2}(\textbf{p}_t^l)  \\
&\textbf{p}_{t+1}^l = \textbf{p}_t^l + \textbf{w}_{1 \to 3}(\textbf{p}_t^l) \\
&\textbf{p}_{t+1}^r = \textbf{p}_t^l + \textbf{w}_{1 \to 4}(\textbf{p}_t^l)
\end{cases}.
\label{eq:pixels}
\end{equation}
A pixel directly moves from $I_1$ to $I_4$ shall be identical to the movement from $I_1$ to $I_2$ and from $I_2$ to $I_4$. That is,
\begin{equation}
\begin{split}
\textbf{w}_{1 \to 4}(\textbf{p}_t^l)
& = (\textbf{p}_{t+1}^r - \textbf{p}_t^r) +  (\textbf{p}_t^r - \textbf{p}_t^l) \\
& = \textbf{w}_{2 \to 4}(\textbf{p}_t^r) + \textbf{w}_{1 \to 2}(\textbf{p}_t^l).
\end{split}
\label{eq:movement124}
\end{equation}
Similarly, if the pixel moves from $I_1$ to $I_3$ and from $I_3$ to $I_4$, then
\begin{equation}
\begin{split}
\textbf{w}_{1 \to 4}(\textbf{p}_t^l)
&= (\textbf{p}_{t+1}^r - \textbf{p}_{t+1}^l) +  (\textbf{p}_{t+1}^l - \textbf{p}_t^l) \\
&=\textbf{w}_{3 \to 4}(\textbf{p}_{t+1}^l) + \textbf{w}_{1 \to 3}(\textbf{p}_t^l).
\end{split}
\label{eq:movement134}
\end{equation}
From Eq.~(\ref{eq:movement124})  and (\ref{eq:movement134}), we obtain,
\begin{equation}
\textbf{w}_{2 \to 4}(\textbf{p}_t^r) - \textbf{w}_{1 \to 3}(\textbf{p}_t^l) = \textbf{w}_{3 \to 4}(\textbf{p}_{t+1}^l) - \textbf{w}_{1 \to 2}(\textbf{p}_t^l).
\label{eq:movement2413}
\end{equation}
For stereo matching, the correspondence pixel shall lie on the epipolar lines. Here, we only consider rectified stereo cases, where epipolar lines are horizontal. Then,  Eq.(\ref{eq:movement2413}) becomes
\begin{equation}
\begin{cases}
&u_{2 \to 4}(\textbf{p}_t^r) - u_{1 \to 3}(\textbf{p}_t^l) = u_{3 \to 4}(\textbf{p}_{t+1}^l) - u_{1 \to 2}(\textbf{p}_t^l) \\
&v_{2 \to 4}(\textbf{p}_t^r) - v_{1 \to 3}(\textbf{p}_t^l) = 0
\end{cases}.
\label{eq:constraints}
\end{equation}
Note Eq.~(\ref{eq:constraints}) is exactly the same as Eq.~(\ref{eq: 09}).

In addition, since epipolar lines are horizontal, we can re-write Eq.~(\ref{eq:movement124}) and (\ref{eq:movement134}) as follows:
\begin{equation}
\begin{cases}
&u_{1 \to 4}(\textbf{p}_t^l) = u_{2 \to 4}(\textbf{p}_t^r) + u_{1 \to 2}(\textbf{p}_t^l) \\
&v_{1 \to 4}(\textbf{p}_t^l) = v_{2 \to 4}(\textbf{p}_t^r) \\
&u_{1 \to 4}(\textbf{p}_t^l) = u_{3 \to 4}(\textbf{p}_{t+1}^l) + u_{1 \to 3}(\textbf{p}_t^l) \\
&v_{1 \to 4}(\textbf{p}_t^l) = v_{1 \to 3}(\textbf{p}_t^l)
\end{cases}.
\label{eq: triangle constraints}
\end{equation}

This leads to the two forms of geometric constraints we used in our training loss functions: quadrilateral constraint   (\ref{eq:constraints}) and triangle constraint (\ref{eq: triangle constraints}).

\begin{table*}[t]
\caption{Quantitative evaluation of optical flow estimation on KITTI. Bold fonts highlight the best results among supervised and unsupervised methods. Parentheses mean that training and testing are performed on the same dataset. \texttt{fg} and \texttt{bg} denote results of foreground and background regions respectively.}
\label{FlowResult}
\centering
\resizebox{1\textwidth}{!}{
\begin{tabular}{ c  l  c  c  c c c c  c c c c c c}
 \toprule
   &   \multirow{3}{*}{Method} & & \multicolumn{6}{c}{KITTI 2012} & \multicolumn{5}{c}{KITTI 2015} \\
   \cmidrule(l{3mm}r{3mm}){4-9}    \cmidrule(l{3mm}r{3mm}){10-14}
   & & Train &\multicolumn{2}{c}{train} &\multicolumn{4}{c}{test} & \multicolumn{2}{c}{train} & \multicolumn{3}{c}{test}  \\
   \cmidrule(l{3mm}r{3mm}){4-5}  \cmidrule(l{3mm}r{3mm}){6-9} \cmidrule(l{3mm}r{3mm}){10-11} \cmidrule(l{3mm}r{3mm}){12-14}
                 &  & Stereo & EPE-all & EPE-noc & EPE-all & EPE-noc & Fl-all & Fl-noc & EPE-all & EPE-noc & Fl-all &Fl-fg & Fl-bg \\
  \midrule
   \multirow{9}{*}{\rotatebox[origin=c]{90}{Supervised}}
   & SpyNet~\cite{ranjan2017optical}    &\xmark      & 3.36  & -- & 4.1 & 2.0 & 20.97\% & 12.31\% & --  & --  & 35.07\% & 43.62\% & 33.36\% \\
   & FlowFieldsCNN~\cite{bailer2017cnn} &\xmark   & --    & -- & 3.0 & 1.2 & 13.01\% &4.89\% & -- & -- & 18.68\% & 20.42\% & 18.33\% \\
   & DCFlow~\cite{XRK2017}              &\xmark      & --    & -- & --  & --  & --      & --   & -- & -- & 14.86\% & 23.70\% & 13.10\% \\
   & FlowNet2~\cite{ilg2017flownet}     &\xmark   & (1.28)& --  & 1.8 & 1.0 & 8.80\% & 4.82\% & (2.3)  & -- & 10.41\% &8.75\% & 10.75\% \\
   & UnFlow-CSS~\cite{Meister:2018:UUL} &\xmark   & (1.14)& (0.66)& 1.7 & 0.9&8.42\% & 4.28\% &(1.86) & -- &11.11\% & 15.93\% &10.15\% \\
   & LiteFlowNet~\cite{hui18liteflownet}&\xmark & (1.05)& -- & 1.6 & \textbf{0.8} & 7.27\% &\textbf{3.27\%} &(1.62) & -- & 9.38\% & 7.99\% & 9.66\% \\
   & PWC-Net~\cite{sun2018pwc}          &\xmark& (1.45)& -- &1.7 & 0.9& 8.10\% &4.22\%&(2.16)&--& 9.60\% &9.31\%&9.66\% \\
   & MFF~\cite{ren2018fusion}           &\xmark & --    & -- & 1.7 & 0.9 &7.87\% &4.19\% & -- & -- & \textbf{7.17\%} &\textbf{7.25\%} &\textbf{7.15\%} \\
   & SelFlow~\cite{Liu:2019:SelFlow}    &\xmark& \textbf{(0.76)} & -- & \textbf{1.5} & 0.9 & \textbf{6.19\%} & 3.32\% & \textbf{(1.18)} & -- & 8.42\% & 7.61\% & 12.48\% \\
  \midrule
   \multirow{12}{*}{\rotatebox[origin=c]{90}{Unsupervised}}
   & BackToBasic~\cite{jason2016back}   &\xmark   & 11.3  & 4.3  & 9.9   & 4.6 & 43.15\% & 34.85\% & --  & -- & -- & -- & --      \\
   & DSTFlow~\cite{ren2017unsupervised} &\xmark   & 10.43 & 3.29 & 12.4  & 4.0 & --  & -- & 16.79 & 6.96 & 39\% & -- & --   \\
   & UnFlow-CSS~\cite{Meister:2018:UUL} &\xmark   & 3.29  & 1.26 & --    & --  & -- & -- & 8.10  & -- & 23.30\% & -- & --\\
   & OccAwareFlow~\cite{wang2018occlusion} &\xmark  & 3.55  & --   & 4.2   & --  & -- & -- & 8.88  & -- & 31.2\% & -- & -- \\
   & MultiFrameOccFlow-None~\cite{Janai2018ECCV} &\xmark & --    & --   & --    & --  & -- & -- & 6.65  & 3.24 & -- & -- & --     \\
   & MultiFrameOccFlow-Soft~\cite{Janai2018ECCV} &\xmark & --    & --   & --    & --  & -- & -- & 6.59  & 3.22 & 22.94\% & -- & --\\
   & DDFlow~\cite{Liu:2019:DDFlow}      &\xmark &2.35   & 1.02 & 3.0   & 1.1 & 8.86\% & 4.57\% & 5.72 & 2.73 & 14.29\% & 20.40\% & 13.08\% \\
   & SelFlow~\cite{Liu:2019:SelFlow}    &\xmark &1.69   & 0.91 & 2.2   & 1.0 & 7.68\% & 4.31\% & 4.84 & 2.40 & 14.19\% & 21.74\% & 12.68\% \\
   & Lai~\etal~\cite{lai2019bridging}   &\cmark      &2.56   & 1.39 & --    & --  & --   & -- & 7.134 & 4.306 & -- & -- & -- \\
   & UnOS~\cite{wang2019unos}           &\cmark      &1.64   & 1.04 &  1.8  & --  & -- & -- & 5.58 & -- & 18.00\% & -- & --\\
   \cline{2-14}
   & Our+$L_p$+$L_q$+$L_t$            &\cmark          &4.91   & 0.84 & --    & --  &--   & -- &7.88  & 2.24 & -- & -- & --\\
   & Ours+$L_p$+$L_q$+$L_t$+Self-Supervision      &\cmark             &\textbf{1.45} &\textbf{0.82}& \textbf{1.7} &\textbf{0.9} &\textbf{7.63\%}&\textbf{4.02\%} & \textbf{3.54} & \textbf{2.12} & \textbf{11.10\%} & \textbf{16.67\%} & \textbf{9.99\%}\\
 \bottomrule \end{tabular} }
\end{table*}

\section{Method}
In this section, we first dig into the bottlenecks of the state-of-the-art two-stage self-supervised learning framework~\cite{Liu:2019:DDFlow,Liu:2019:SelFlow}. Then we describe an enhanced proxy learning approach, which can improve its performance greatly in both two stages.

\subsection{Two-Stage Self-Supervised Learning Scheme}
Both DDFlow~\cite{Liu:2019:DDFlow} and SelFlow~\cite{Liu:2019:SelFlow} employ a two-stage learning approaches to learning optical flow in a self-supervised manner. In the first stage, they train a teacher model to estimate optical flow for \texttt{non-occluded} pixels. In the second stage, they first pre-process the input images, \eg, cropping and inject superpixel noises to create hand-crafted   \texttt{occlusions}, then the predictions of teacher model for those \texttt{non-occluded} pixels are regarded as ground truth to guide a student model to learn optical flow of hand-crafted \texttt{occluded} pixels.

The general pipeline is reasonable, but the definition of \texttt{occlusion} is in a heuristic manner. At the first stage, forward-backward consistency is employed to detect whether the pixel is \texttt{occluded}. However, this brings in errors  because many pixels are still \texttt{non-occluded} even they violate this principle, and vice versa. Instead, it would be more proper to call those pixels reliable or confident if they pass the forward-backward consistency check. From this point of view, creating hand-crafted \texttt{occlusions} can be regard as creating more challenging conditions, under which the prediction would be less confident. Then in the second stage, the key point is to let confident predictions to supervise those less confident predictions.

During the self-supervised learning stage, the student model is able to handle more challenging conditions. As a result, its performance improves  not only for those occluded pixels, but also for non-occluded pixels. Because when creating challenging conditions, both occluded regions and non-occluded regions become more challenging. The reason why optical flow for occluded pixels improves more than non-occluded regions is that, during the first stage, photometric loss does not hold for occluded pixels, the model just does not have the ability to predict them. In the second stage, the model has the ability to learn optical flow of occluded pixels for the first time, therefore its performance improves a lot.
To lift the upper bound of confident predictions, we propose to utilize stereoscopic videos  to reveal their geometric nature.

\begin{figure*}[th]
\centering
\begin{tabular}{@{}c@{\hspace{0.5mm}}c@{\hspace{0.5mm}}c@{\hspace{0.5mm}}c@{}}
\begin{overpic} [width=0.245\textwidth]{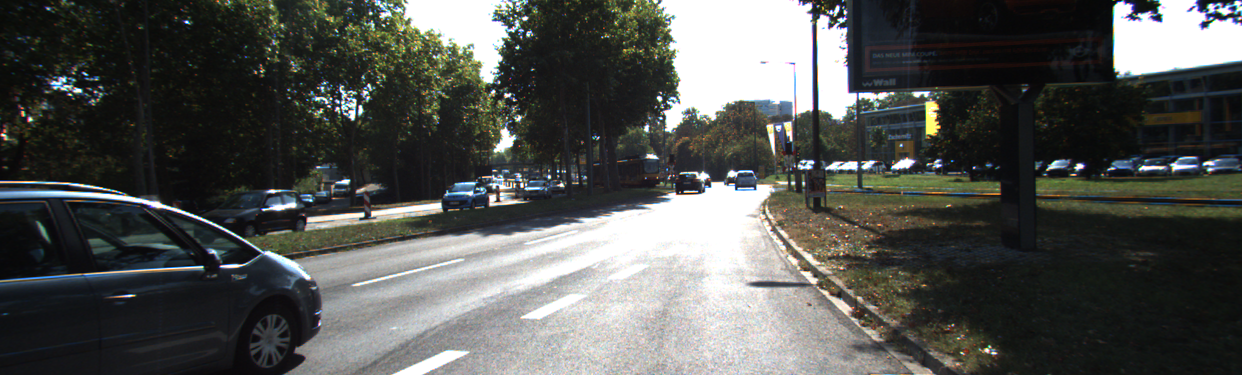} \put(4, 20){\color{red} \textbf{$I_l$}} \end{overpic} &
\includegraphics[width=0.245\textwidth]{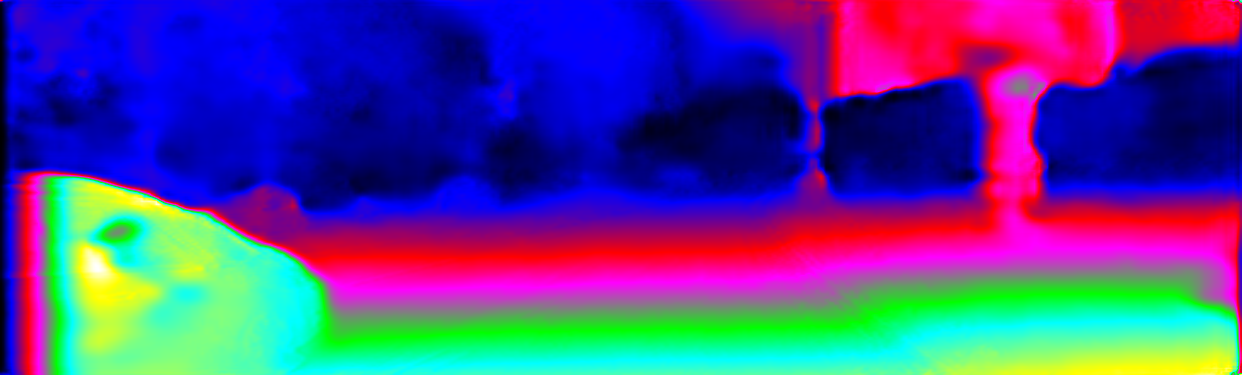} &
\includegraphics[width=0.245\textwidth]{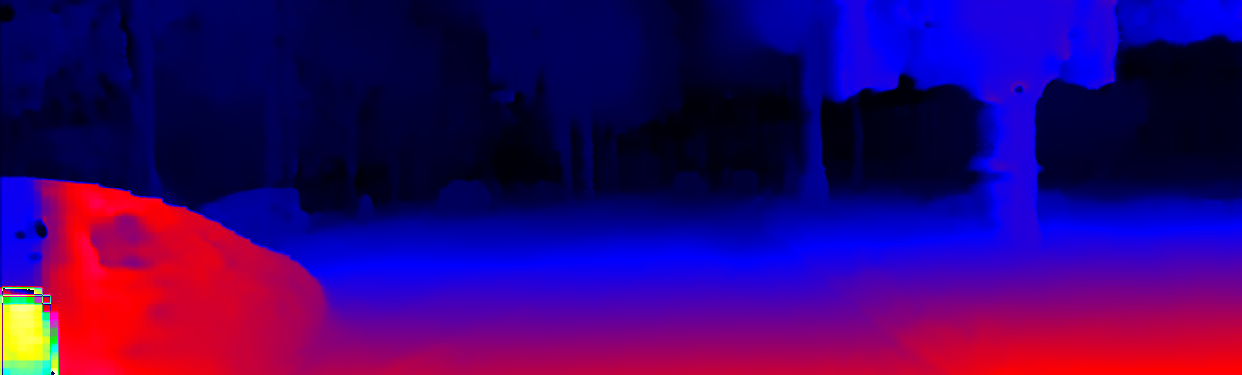} &
\includegraphics[width=0.245\textwidth]{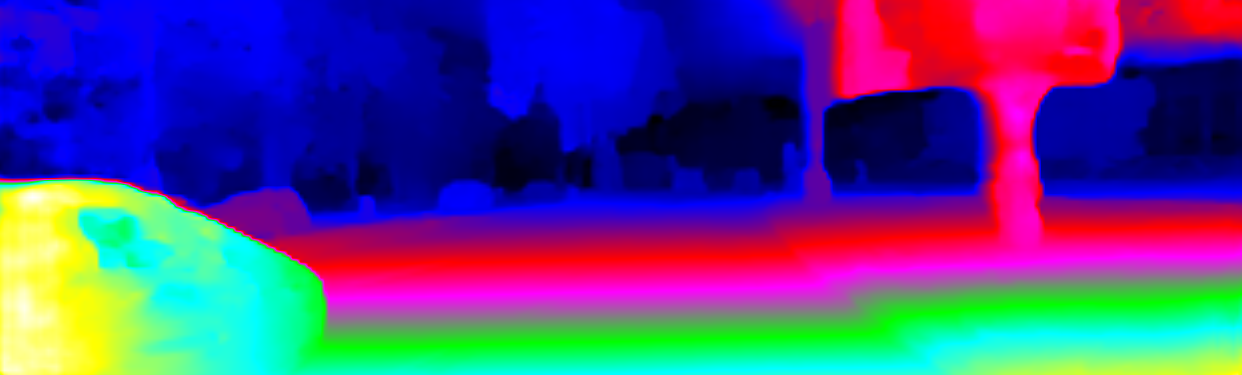} \\
\begin{overpic} [width=0.245\textwidth]{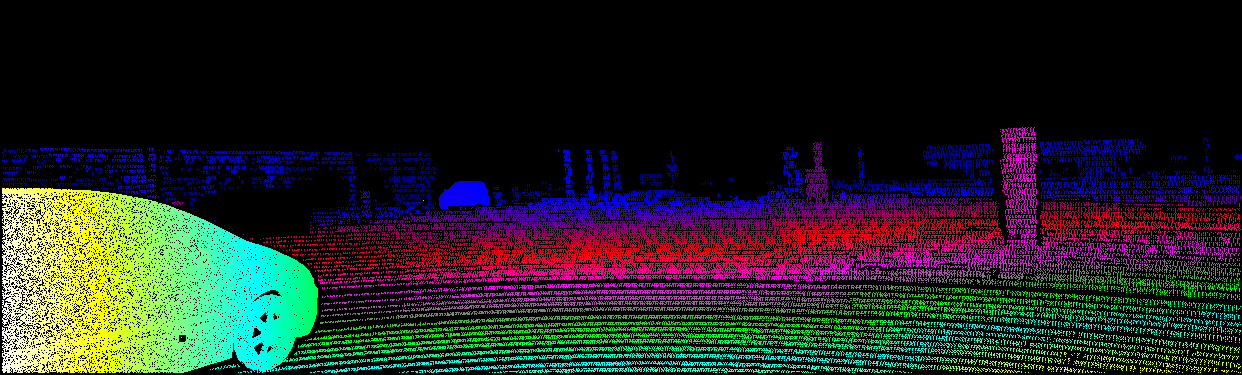} \put(4, 20){\color{red} \textbf{GT Disparity}} \end{overpic}&
\begin{overpic} [width=0.245\textwidth]{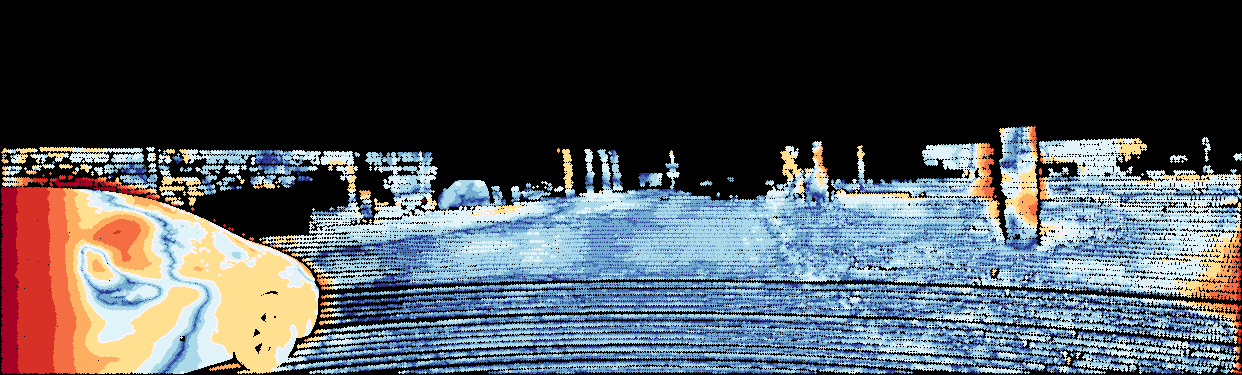} \put(4, 20){\color{white} \textbf{D1: 6.22\%}} \end{overpic}&
\begin{overpic} [width=0.245\textwidth]{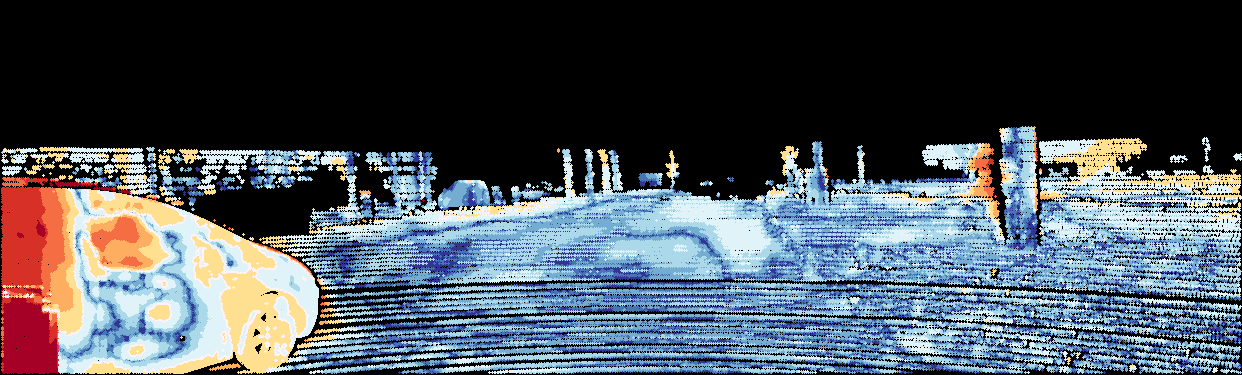} \put(4, 20){\color{white} \textbf{D1: 5.23\%}} \end{overpic}&
\begin{overpic} [width=0.245\textwidth]{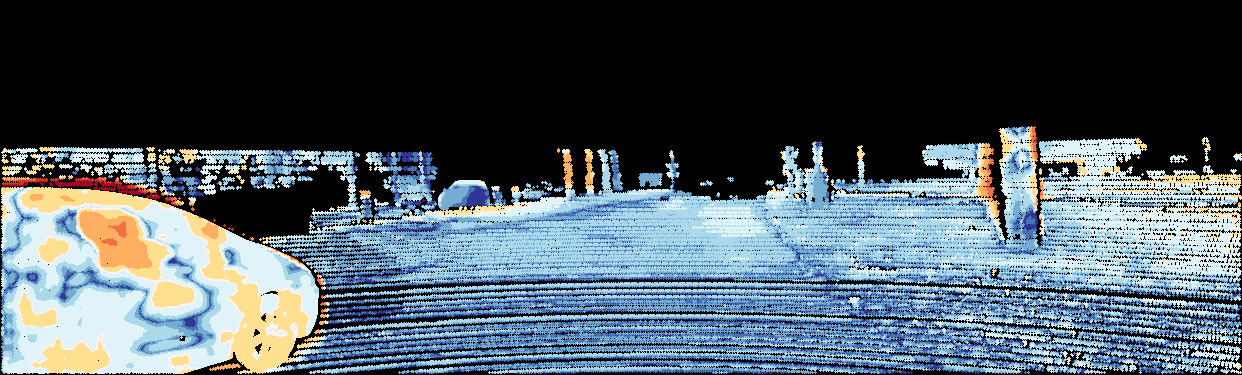}\put(4, 20){\color{white} \textbf{D1: 1.56\%}} \end{overpic} \\

\begin{overpic} [width=0.245\textwidth]{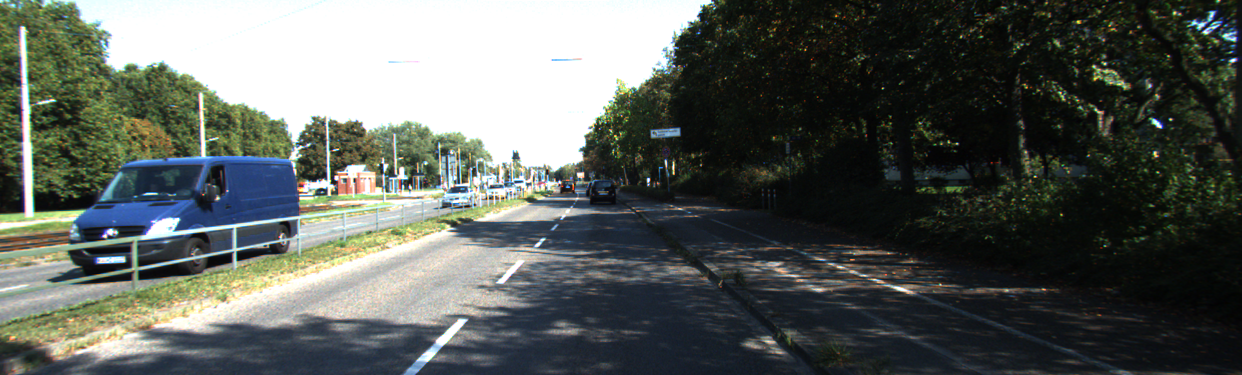} \put(4, 20){\color{red} \textbf{$I_l$}} \end{overpic} &
\includegraphics[width=0.245\textwidth]{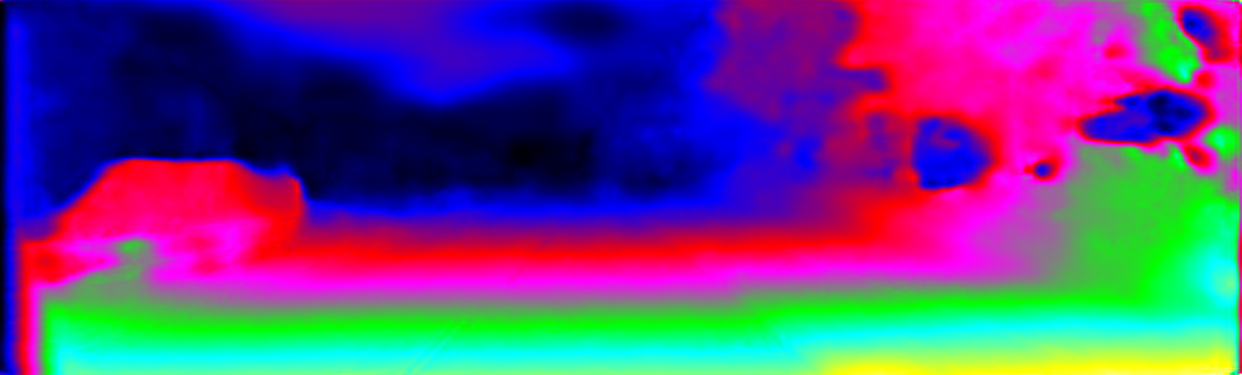} &
\includegraphics[width=0.245\textwidth]{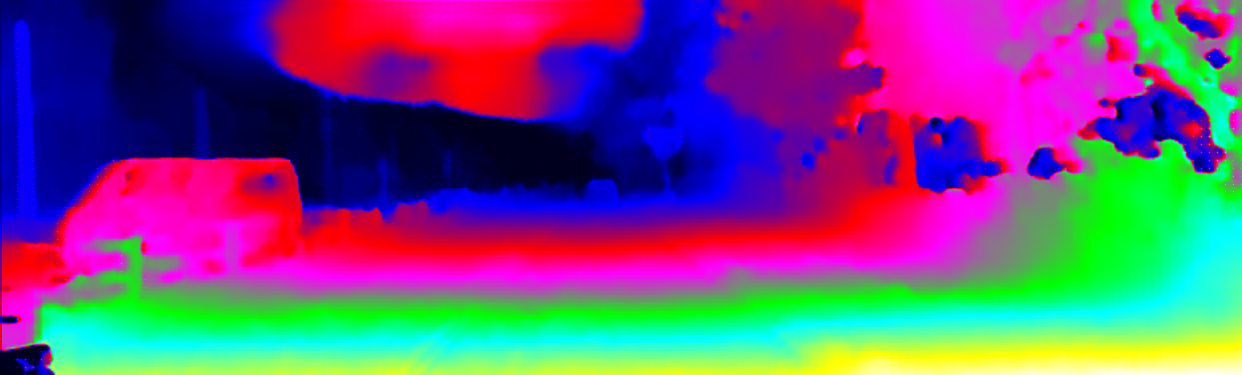} &
\includegraphics[width=0.245\textwidth]{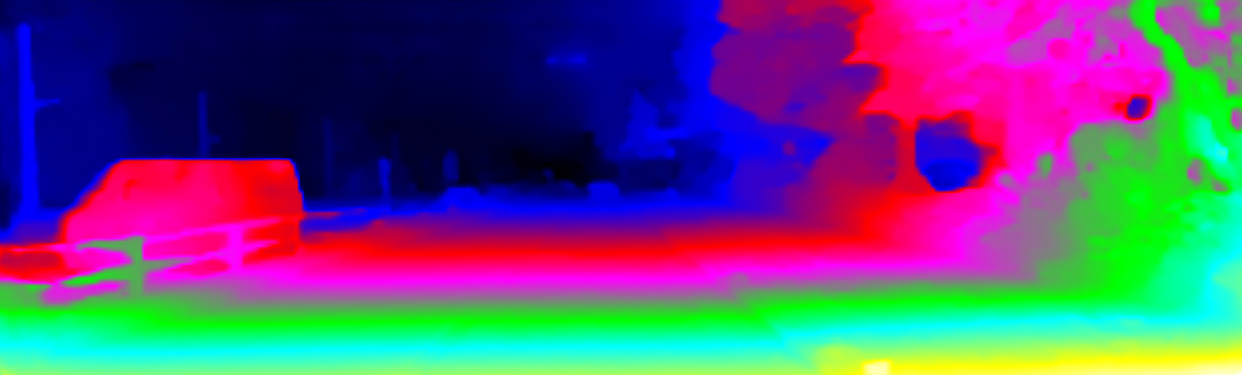} \\
\begin{overpic} [width=0.245\textwidth]{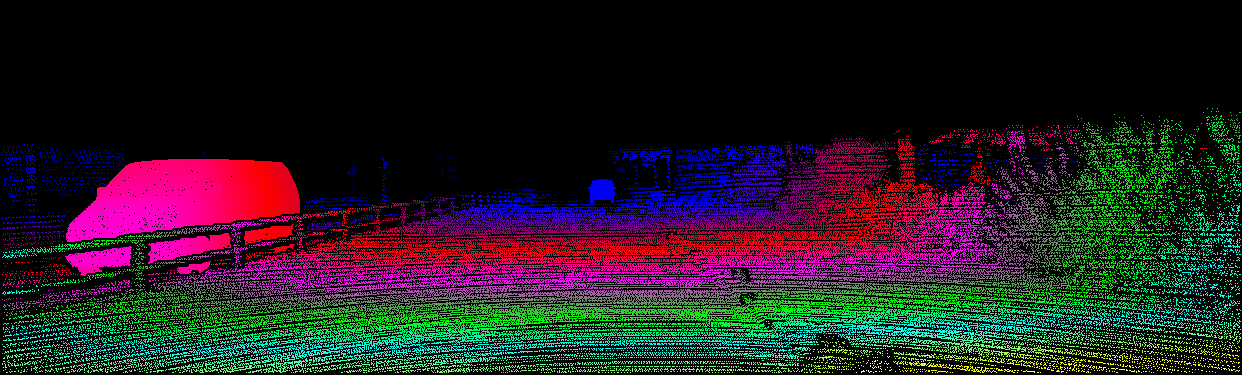} \put(4, 20){\color{red} \textbf{GT Disparity}} \end{overpic}&
\begin{overpic} [width=0.245\textwidth]{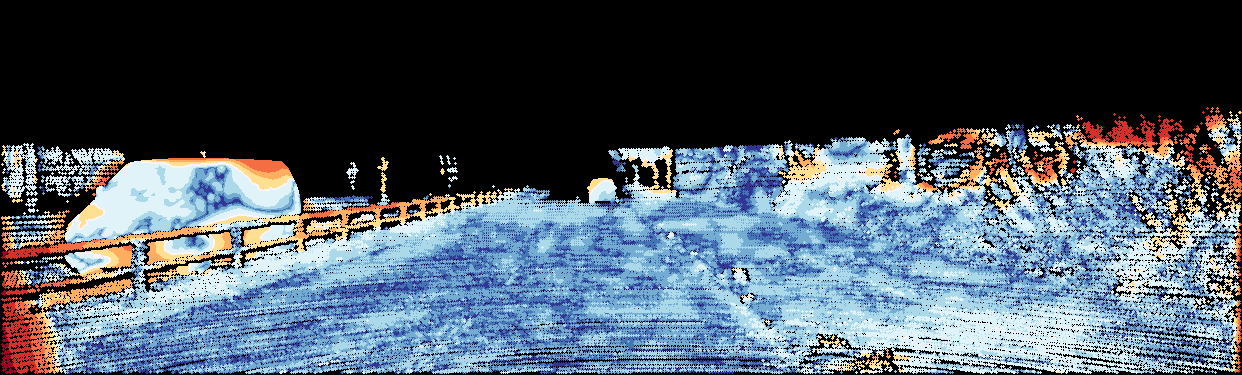} \put(4, 20){\color{white} \textbf{D1: 9.63\%}} \end{overpic}&
\begin{overpic} [width=0.245\textwidth]{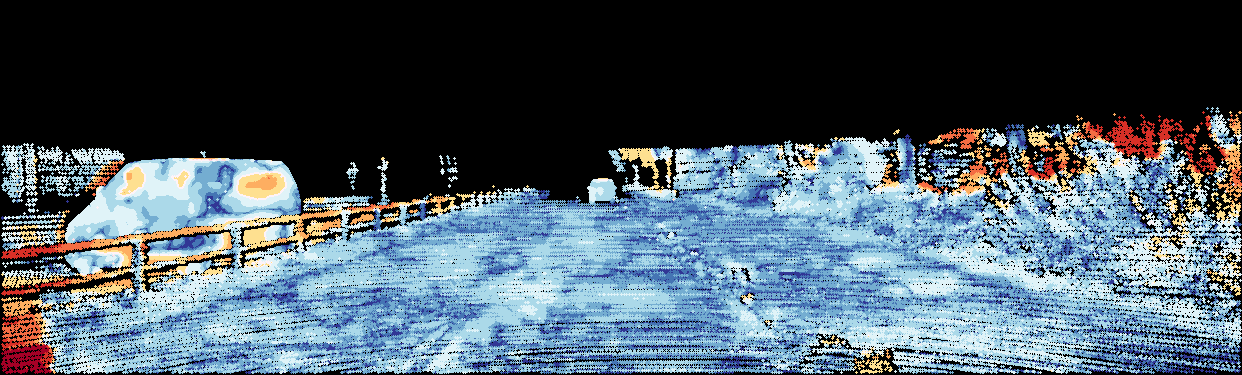} \put(4, 20){\color{white} \textbf{D1: 7.55\%}} \end{overpic}&
\begin{overpic} [width=0.245\textwidth]{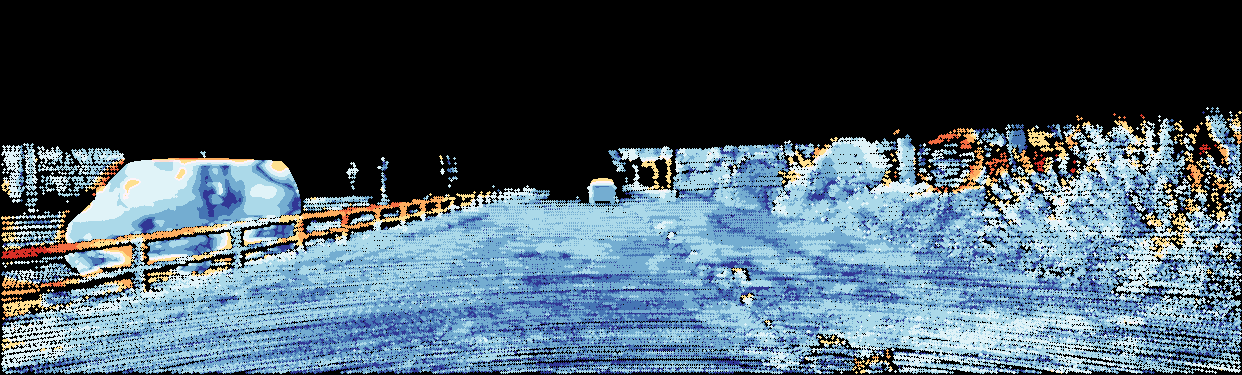}\put(4, 20){\color{white} \textbf{D1: 4.38\% }} \end{overpic} \\
(a) $I_l$ and GT Disparity & (b) Godard~\etal~\cite{Godard_2017_CVPR}  & (c) SeqStereo~\cite{yang2018segstereo}  & (d) Ours \\
  \end{tabular}
\caption{Qualitative evaluation with other unsupervised stereo matching methods on KITTI 2015 training dataset. For each case, the top row is stereo disparity and the bottom row is error map. Our models estimate more accurate disparity maps (\eg, image boundary regions and moving-object boundary regions). Lower D1 is better.}
\label{ComparisonDisp}
\end{figure*}

\subsection{Proxy Learning Scheme}
Following~\cite{Liu:2019:DDFlow,Liu:2019:SelFlow}, our proxy learning scheme contains two stages and our network structure is built upon PWC-Net~\cite{sun2018pwc}.

\mypara{Stage 1: Predicting confident optical flow with geometric constraints.}  With the estimated  optical flow map $\textbf{w}_{i \to j}$, we  warp the  target image $I_j$ toward the reference image $I_i$. Then we  measure the difference between the warped target image  $I_{j \to i} ^ w$ and the reference image $I_i$ with a photometric loss. Similar to \cite{Meister:2018:UUL,Liu:2019:DDFlow,Liu:2019:SelFlow}, we employ forward-backward  consistency check to compute a confident map, where value 1 indicates the prediction is confident, 0 indicates the prediction is non-confident.

Apart from photometric loss, we also apply geometric constraints to our teacher model, including the triangle constraint and quadrilateral constraint. Note that geometric constraints are only applied to those confident pixels. This turns out to highly effective and greatly improves the accuracy of those confident predictions.

\mypara{Stage 2: Self-supervised learning from teacher model to student model.} As discussed earlier, the key point of self-supervision is to create challenging input-output pairs. In our framework, we create challenging conditions by random cropping input image pairs,   injecting random noise into the second image, random scale (down-sample) the input image pairs, to make correspondence learning more difficult. These hard input-output pairs push the network to capture  more information, resulting in a large performance gain in practice.

Different from \cite{Liu:2019:DDFlow,Liu:2019:SelFlow}, we do not  distinguish between “occluded” and “non-occluded” pixels anymore in the self-supervision stage. 
As forward-backward consistency check cannot perfectly determine whether a pixel is occluded,   there  may be many erroneous judgments. In this case, the confident prediction from teacher model will provide guidance for both occluded or non-occluded pixels no matter forward-backward check is employed or not.

Next, we describe our training losses for each stage.

\begin{table*}[t]
\caption{Quantitative evaluation of stereo disparity on KITTI training datasets (apart from the last columns). Our single model achieves the highest accuracy among all unsupervised stereo learning methods. \texttt{*} denotes that we use their pre-trained model to compute the numbers, while other numbers are from their paper. Note that Guo \textit{et al.}~\cite{Guo_2018_ECCV}  pre-train stereo model on synthetic Scene Flow dataset with ground truth disparity before fine-tuning on KITTI dataset.
}
\label{StereoResult}
\centering
\resizebox{\textwidth}{!}{
\begin{tabular}{ l  c c c c c c c c c c c c }
 \toprule
   \multirow{2}{*}{Method} & \multicolumn{6}{c}{KITTI 2012} & \multicolumn{6}{c}{KITTI 2015} \\
    \cmidrule(l{3mm}r{3mm}){2-7}    \cmidrule(l{3mm}r{3mm}){8-13}
   &  EPE-all & EPE-noc & EPE-occ & D1-all & D1-noc & D1-all (test) & EPE-all & EPE-noc & EPE-occ & D1-all & D1-noc & D1-all (test) \\
   \midrule
   Joung \textit{et al.}~\cite{joung2019unsupervised} & -- & -- & -- & -- & -- & 13.88\% & -- & -- & -- & 13.92\% & -- & --\\
   Godard  \textit{et al.}~\cite{Godard_2017_CVPR} * & 2.12 & 1.44 & 30.91& 10.41\% & 8.33\% & -- & 1.96 & 1.53 & 24.66 & 10.86\% & 9.22\% & -- \\
   Zhou \textit{et al.}~\cite{Zhou_2017_ICCV}  & -- & -- & -- & -- & -- & -- & -- & -- & -- & 9.41\% & 8.35\% & --\\
   OASM-Net ~\cite{li2018occlusion} & -- & -- & -- & 8.79\% & 6.69\% & 8.60\% & -- & -- & -- & -- & -- & 8.98\%\\
   SeqStereo \textit{et al.}~\cite{yang2018segstereo} * & 2.37 & 1.63 & 33.62 & 9.64\% & 7.89\% & -- & 1.84 & 1.46 & 26.07 & 8.79\% & 7.7\% & -- \\
   Liu  \textit{et al.}~\cite{liu2019unsupervised} * & 1.78 & 1.68 & 6.25 & 11.57\% & 10.61\% & -- & 1.52 & 1.48 & 4.23 & 9.57\% & 9.10\% & --\\
   Guo \textit{et al.}~\cite{Guo_2018_ECCV} * & 1.16 & 1.09 & 4.14 & 6.45\% & 5.82\% & -- & 1.71 & 1.67 & 4.06 & 7.06\% & 6.75\% & --\\
   UnOS~\cite{wang2019unos}  & -- & -- & -- & -- & -- & 5.93\% & -- & -- & -- & \textbf{5.94\%} & -- & 6.67\%\\
   \midrule
   Ours+$L_p$ & 1.73 & 1.13 & 27.03 & 7.88\% & 5.87\% & -- & 1.79 & 1.40 & 25.24 & 9.83\% & 7.74\% & --\\
   Ours+$L_p$+$L_q$+$L_t$ &  1.62  & 0.94 & 29.26 & 6.69\% & 4.69\% & -- &1.67  & \textbf{1.31} & 19.55 & 8.62\% & 7.15\% & -- \\
   Ours+$L_p$+$L_q$+$L_t$+Self-Supervision & \textbf{1.01}  & \textbf{0.93} & \textbf{4.52}  & \textbf{5.14\%} & \textbf{4.59\%} & \textbf{5.11\%} & \textbf{1.34} & \textbf{1.31} & \textbf{2.56} & 6.13\% & \textbf{5.93\%} & \textbf{6.61\%} \\
\bottomrule \end{tabular} }
\end{table*}

\subsection{Loss Functions}
For stage 1, our loss function mainly contains three parts: photometric loss $L_p$, triangle constraint loss $L_t$ and quadrilateral constraint loss $L_q$. For stage 2, we only apply self-supervision loss $L_s$.

\mypara{Photometric loss.} Photometric loss is based on the brightness consistency assumption, which only works for non-occluded pixels. During our experiments, we employ census transform, which has shown to be robust for illumination change~\cite{Meister:2018:UUL,Liu:2019:DDFlow,Liu:2019:SelFlow}. Denote $M_{i \to j}$  as the confident map from $I_i$ to $I_j$ is $M_{i \to j}$, then $L_p$  is defined as,
\begin{equation}
  L_p = \sum_{i, j} \frac {\sum_{\textbf{p}}\psi{(I_i(\textbf{p}) - I_{j \to i}^{w}(\textbf{p}))} \odot M_{i \to j}(\textbf{p})} {\sum_{\textbf{p}} M_{i \to j}(\textbf{p})}
  \label{eq:photometric_loss},
\end{equation}
where $\psi(x) = (|x| + \epsilon)^q$. During our experiments, we set $\epsilon = 0.01$ and $q = 0.4$.

\mypara{Quadrilateral constraint loss.} Quadrilateral constraint describes the geometric relationship between optical flow and stereo disparity. Here, we only employ $L_q$ to those confident pixels. Take  $\textbf{w}_{1 \to 4}$, $\textbf{w}_{2 \to 4}$, $\textbf{w}_{1 \to 2}$ and $\textbf{w}_{3 \to 4}$ for an example, we   first compute the confident map for quadrilateral constraint $M_q(\textbf{p}) = M_{1 \to 2}(\textbf{p}) \odot M_{1 \to 3}(\textbf{p})  \odot M_{1 \to 4}(\textbf{p})$. Then according to Eq.~(\ref{eq:constraints}),  we   divide $L_q$ into two components on the $x$ direction $L_{qu}$ and $y$ direction $L_{qv}$ respectively:
\begin{equation}
\begin{aligned}
L_{qu} &= \sum_{\textbf{p}_t^{l}}  \psi(u_{1 \to 2}(\textbf{p}_t^l) + u_{2 \to 4}(\textbf{p}_t^r) - u_{1 \to 3}(\textbf{p}_t^l) - \\
&  u_{3 \to 4}(\textbf{p}_{t+1}^l)) \odot M_q(\textbf{p}_t^l) / \sum_{\textbf{p}_t^l} M_q(\textbf{p}_t^l)
\end{aligned}
\end{equation}
\begin{equation}
L_{qv} = \sum_{\textbf{p}_t^{l}} \psi(v_{2 \to 4}(\textbf{p}_t^r) - v_{1 \to 3}(\textbf{p}_t^l)) \odot M_q(\textbf{p}_t^l) / \sum_{\textbf{p}_t^l} M_q(\textbf{p}_t^l)
\end{equation}
where $L_q = L_{qu} + L_{qv}$. Quadrilateral constraint loss at other directions are computed in the same way.

\mypara{Triangle constraint loss.} Triangle constraint describes the relationship between optical flow, stereo disparity and cross-view optical flow. Similar to quadrilateral constraint loss, we only employ $L_t$ to confident pixels. Take $\textbf{w}_{1 \to 3}$, $\textbf{w}_{2 \to 4}$, $\textbf{w}_{1 \to 2}$ as an example, we   first compute the confident map for triangle constraint $M_t(\textbf{p}) = M_{1 \to 2}(\textbf{p}) \odot M_{1 \to 4}(\textbf{p})$, then according to Eq.~(\ref{eq:movement124}), $L_t$ is defined as follows,
\begin{equation}
L_{tu} = \frac {\sum_{\textbf{p}_t^{l}} \psi(u_{1 \to 4}(\textbf{p}_t^l) - u_{2 \to 4}(\textbf{p}_t^r) - u_{1 \to 2}(\textbf{p}_t^l)) \odot M_t(\textbf{p})}  {\sum_{\textbf{p}_t^l} M_t(\textbf{p}_t^l)}，
\end{equation}
\begin{equation}
L_{tv} = \sum_{\textbf{p}_t^{l}} \psi(v_{1 \to 4}(\textbf{p}_t^l) - v_{2 \to 4}(\textbf{p}_t^r)) \odot M_t(\textbf{p})  {\sum_{\textbf{p}_t^l} M_t(\textbf{p}_t^l)},
\end{equation}
where $L_t = L_{tu} + L_{tv}$. Triangle constraint losses at other directions are computed in the same way.

The final loss function for teacher model is $L = L_p + \lambda_1 L_q + \lambda_2 L_t$, where we set $\lambda_1=0.1$ and $\lambda_2$ = 0.2 during experiments.

\textbf{Self-Supervision loss.} During the first stage, we train our teacher model to compute proxy optical flow $\textbf{w}$ and confident map $M$, then we  define our self-supervision loss as,
\begin{equation}
L_s = \sum_{i, j} \frac {\sum_{\textbf{p}} \psi(\textbf{w}_{i \to j}(\textbf{p}) - \widetilde{\textbf{w}}_{i \to j}(\textbf{p})) \odot M_{i \to j}(\textbf{p})} {\sum_{\textbf{p}} M_{i \to j}(\textbf{p})}.
\end{equation}
At test time, only the student model is needed, and we can use it to estimate both optical flow and stereo disparity.

\section{Experiments}
We evaluate our method on the challenging KITTI 2012 and KITTI 2015 datasets and compare our method with state-of-the-art unsupervised and supervised optical flow learning methods. Besides, since our method is able to predict stereo disparity, we also compare its stereo matching performance with related methods. we make our code and models publicly available at \texttt{https://github.com/ppliuboy/Flow2Stereo}

\subsection{Experimental Setting}
During training, we use the raw multi-view extensions of KITTI 2012~\cite{geiger2012we} and KITTI 2015~\cite{menze2015object} and exclude neighboring frames (frame 9-12) as~\cite{ren2017unsupervised,wang2018occlusion,Liu:2019:DDFlow,Liu:2019:SelFlow}. For evaluation, we use the training sets of KITTI 2012 and KITTI 2015 with ground truth optical flow and disparity. We also submit our results to optical flow and stereo matching benchmarks for comparison with current state-of-the-art methods.

\begin{table*}[h]
\caption{Ablation study on KITTI training datasets. For self-supervision, \texttt{v1} means employing self-supervision of \cite{Liu:2019:DDFlow,Liu:2019:SelFlow}, \texttt{v2} means not distinguishing between occluded and non-occluded pixels, \texttt{v3} means adding more challenging conditions (our final model), and \texttt{v4} means adding geometric constraints in the self-supervision stage (slightly degrade the performance).}
\label{AblationStudy}
\centering
\resizebox{\textwidth}{!}{
\begin{tabular}{ c c c c c c c c c c c c c cccc }
   \toprule
    \multirow{2}{*}{$L_p$}    &\multirow{2}{*}{$L_q$} & \multirow{2}{*}{$L_t$} & \multicolumn{4}{c}{Self-Supervision} &\multicolumn{5}{c}{KITTI 2012}& \multicolumn{5}{c}{KITTI 2015}\\
    \cmidrule(l{3mm}r{3mm}){4-7} \cmidrule(l{3mm}r{3mm}){8-12}      \cmidrule(l{3mm}r{3mm}){13-17}
        & & & v1& v2 & v3 & v4 & EPE-all  &  {EPE-noc}   & EPE-occ &  Fl-all  &   {Fl-noc}    &   EPE-all &   {EPE-noc}  &  EPE-occ  &   Fl-all  &   {Fl-noc} \\
    \midrule
  \cmark &\xmark & \xmark & \xmark &\xmark &\xmark   & \xmark&4.41 & 1.06 & 26.54 & 14.18\% & 5.13\%  & 8.20 & 2.85 & 42.01 & 19.50\% & 9.97\%  \\
  \cmark &\cmark & \xmark & \xmark & \xmark &\xmark  & \xmark&5.15 & 0.84 & 33.74 & 13.53\% & 3.42\%  & 8.24 & 2.33 & 45.46 & 18.31\% & 8.15\%  \\
  \cmark &\xmark & \cmark & \xmark & \xmark &\xmark  & \xmark&4.98 & 0.86 & 32.33 & 12.64\% & 3.54\%  & 7.99 & 2.34 & 43.50 & 17.89\% & 8.14\%  \\
  \cmark &\cmark & \cmark & \xmark & \xmark &\xmark  & \xmark&4.91 & 0.84 & 31.81 & 12.57\% & 3.47\%  & 7.88 & 2.24 & 43.92 & 17.68\% & 7.97\%  \\
  \cmark &\xmark & \xmark & \cmark & \xmark& \xmark  & \xmark&1.92 & 0.95 &  7.86 & 6.56\%  & 3.82\%   & 5.85 & 2.96 & 24.17 & 13.26\%  & 9.06\%  \\
  \cmark &\xmark & \xmark & \xmark & \cmark& \xmark  & \xmark&1.89 & 0.93 &  7.76 & 6.44\%  & 3.76\%   & 5.48 & 2.78 & 22.05 & 12.62\%  & 8.53\%  \\
  \cmark &\xmark & \xmark & \xmark & \xmark& \cmark  & \xmark&1.62 & 0.89 & 6.21  & 5.62\%  & 3.38\%  & 4.12 & 2.36 & 15.04 & 10.93\%  & 8.31\%  \\
  \cmark &\cmark & \cmark & \xmark & \xmark & \cmark & \xmark&\textbf{1.45} & \textbf{0.82} & \textbf{5.52} & \textbf{5.29\%} & \textbf{3.27\%} & \textbf{3.54} & \textbf{2.12} & \textbf{12.58} & \textbf{10.04\%}   &\textbf{7.57\%} \\
  \cmark &\cmark & \cmark & \xmark & \xmark & \xmark & \cmark&1.56 & 0.86 & 6.20 & 5.83\% & 3.41\% & 3.66 & 2.16 & 13.18 & 10.44\% & 7.80\%\\
   \bottomrule
  \end{tabular} }
\end{table*}

We implement our algorithm using TensorFlow with Adam optimizer. For teacher model, we set batch size to be 1, since there are 12 optical flow estimations for the 4 images. For student model, batch size is 4. We adopt similar data augmentation strategy as \cite{dosovitskiy2015flownet}. During training, we random crop [320, 896] as input, while during testing, we resize images to resolution [384, 1280]. We employ a two-stage training procedure as \cite{Liu:2019:DDFlow,Liu:2019:SelFlow}. The key difference is that during the first stage, we add geometric constraints which enable our model to predict more accurate reliable predictions. Besides, during the second stage, we do not distinguish between occluded and non-occluded pixels, and set all our confident predictions as ground truth. For each experiment, we set initial learning rate to be 1e-4 and decay it by half every 50k iterations.

For evaluation metrics, we use the standard EPE (average end-point error) and Fl (percentage or erroneous pixels).  A pixel is considered as correctly estimated if end-point error is $<$3 pixel or $<$5\%. For stereo matching, there is another metric $D1$, which shares the same definition as Fl.

\subsection{Main Results}
Our method achieves the best unsupervised results for all evaluation metrics on both KITTI 2012 and KITTI 2015 datasets. More notably, our unsupervised results are even comparable with state-of-the-art supervised learning methods. Our approach bridges the performance gap between supervised learning and unsupervised learning methods for optical flow estimation.

\mypara{Optical Flow.} As shown in Tab.~\ref{FlowResult}, our method outperforms all unsupervised learning method for all metrics on both KITTI 2012 and KITTI 2015 datasets. Specially, on KITTI 2012 dataset, we achieve EPE-all = 1.45 pixels, which achieves 14.2\% relative improvement than previous best SelFLow~\cite{Liu:2019:SelFlow}. For testing set, we achieve EPE = 1.7 pixels, resulting in 22.7\% improvement. More notably, we achieve FL-all = 7.68\% and Fl-noc = 4.02\%, which is even better than state-of-the-art fully supervised learning methods including PWC-Net~\cite{sun2018pwc}, MFF~\cite{ren2018fusion}, and is highly competitive with LiteFlowNet~\cite{hui18liteflownet} and SelFlow~\cite{Liu:2019:SelFlow}.

On KITTI 2015, the improvement is also impressive. For the training set, we achieve EPE-all = 3.54 pixels, resulting in 26.9\% relative improvement than previous best method SelFlow. On the testing benchmark, we achieve Fl-all = 11.10\%, which is not only better than previous best unsupervised learning methods by a large margin (21.8\% relative improvement), but also competitive with state-of-the-art supervised learning methods. To the best of our knowledge, this is the first time that an unsupervised method achieves comparable performance compared with state-of-the-art fully supervised learning methods. Qualitative comparisons with other methods on KITTI 2015 optical flow benchmark are shown in Fig.~\ref{BenchmarkFlow}.

\mypara{Stereo Matching.} We directly apply our optical flow model to stereo matching (only keeping the horizontal direction of flow), it achieves state-of-the-art unsupervised stereo matching performance as shown in Tab.~\ref{StereoResult}. Specially, we reduce EPE-all from 1.61 pixels to 1.01 pixels on KITTI 2012 training dataset and from 1.71 pixels to 1.34 pixels on KITTI 2015 dataset.

Compared with previous state-of-the-art method UnOS~\cite{wang2019unos}, we reduce Fl-all from 5.93\% to 5.11\% on KITTI 2012 testing dataset and from 6.67\% to 6.61\% on KITTI 2015 testing dataset. This is a surprisingly impressive result, since our optical flow model performs even better than other models specially designed for stereo matching. It also demonstrates the generalization capability of our optical flow model toward stereo matching. Qualitative comparisons with other unsupervised stereo matching approaches are shown in Fig.~\ref{ComparisonDisp}.

\subsection{Ablation Study}
We conduct a thorough analysis for different components of our proposed method.

\mypara{Quadrilateral and Triangle Constraints.} We add both constraints during our training in the first stage, aiming to improve the accuracy of confident pixel, since only these confident pixels are used for   self-supervised training in the second stage. confident pixels  are usually non-occluded in the first stage,  because we optimize our model with photometric loss, which only holds for non-occluded pixels.  Therefore, we are concerned about the performance over those non-occluded pixels (not for all pixels).
As shown in the first 4 rows of Tab.~\ref{AblationStudy}, both constraints significantly improve the performance over those non-occluded pixels, and the combination of them produces the best results, while the \texttt{EPE-occ}  may degrade. This is because we are concerned about the performance over those non-occluded pixels, since only confident pixels are used for self-supervised training. Specially, EPE-noc decreases from 1.06 pixels to 0.84 pixels on KITTI 2012 and from 2.85 pixels to 2.24 pixels on KITTI 2015. It is because that we achieve more accurate confident flow predictions, we are able to achieve much better results in the second self-supervision stage. We also achieve big improvement for stereo matching performance over non-occluded pixels as in Tab.~\ref{StereoResult}.

\mypara{Self-Supervision.} We employ four types of self-supervision (check comparison of row 5, 6, 7, 8 in Tab.~\ref{AblationStudy}). For row 5 and row 6 (\texttt{v1} and \texttt{v2}), we show that it does not make much difference to distinguish occluded or non-occluded pixels denoted by forward-backward consistency check. Because forward-backward consistency predicts confident or non-confident flow predictions, but not occluded or non-occluded pixels. Therefore, the self-supervision will be employed to both occluded and non-occluded pixels whenever forward-backward check is employed. Comparing row 6 and row 7 (\texttt{v2} and \texttt{v3}), we show that after adding additional challenging conditions, flow estimation performance is improved greatly. Currently, we are not able to successfully apply geometric constraints in the self-supervision stage. As shown in row 7 and row 8 (\texttt{v2} and \texttt{v3}), geometric constraints will slightly degrade the performance. This is mainly because there is a correspondence ambiguity within occluded pixels, and it is challenging for our geometric consistency to hold for all occluded pixels.

\section{Conclusion}
We have presented a method to jointly learning optical flow and stereo matching with one single  model. We show that geometric constraints improve the quality of those confident predictions, which further help in the self-supervision stage to achieve much better performance. Besides, after digging into the self-supervised learning approaches, we show that creating challenging conditions is the key to improve the performance. Our approach has achieved the best unsupervised optical flow performance on KITTI 2012 and KITTI 2015, and our unsupervised performance is comparable with  state-of-the-art supervised learning methods. More notably, our unified model also achieves state-of-the-art unsupervised stereo matching performance, demonstrating the generalization capability of our model.

\section*{Acknowledgment}
This work was partially supported by the Research Grants Council of the Hong Kong Special Administrative Region, China (RGC C5026-18GF and No. CUHK 14210717 of the General Research Fund).

\clearpage

{\small
\bibliographystyle{ieee_fullname}
\bibliography{egbib}
}
\balance

\clearpage
\twocolumn[
  \begin{@twocolumnfalse}
{
   \newpage
   \null
   \vskip .375in
   \begin{center}
      {\Large \bf Supplementary Material \par}
      \vspace*{24pt}
      {
      \large
      \lineskip .5em
      \begin{tabular}[t]{c}
          
      \end{tabular}
      \par
      }
      \vskip .5em
      \vspace*{12pt}
   \end{center}
}
  \end{@twocolumnfalse}
]
\setcounter{section}{0}
\setcounter{figure}{0}
\setcounter{table}{0}
\setcounter{footnote}{0}
\renewcommand*{\theHsection}{A\thesection}
\renewcommand*{\theHfigure}{A\thefigure}
\renewcommand*{\theHtable}{A\thetable}

\section{Overview}
\begin{figure*}[t]
\centering
\includegraphics[width=0.97\textwidth]{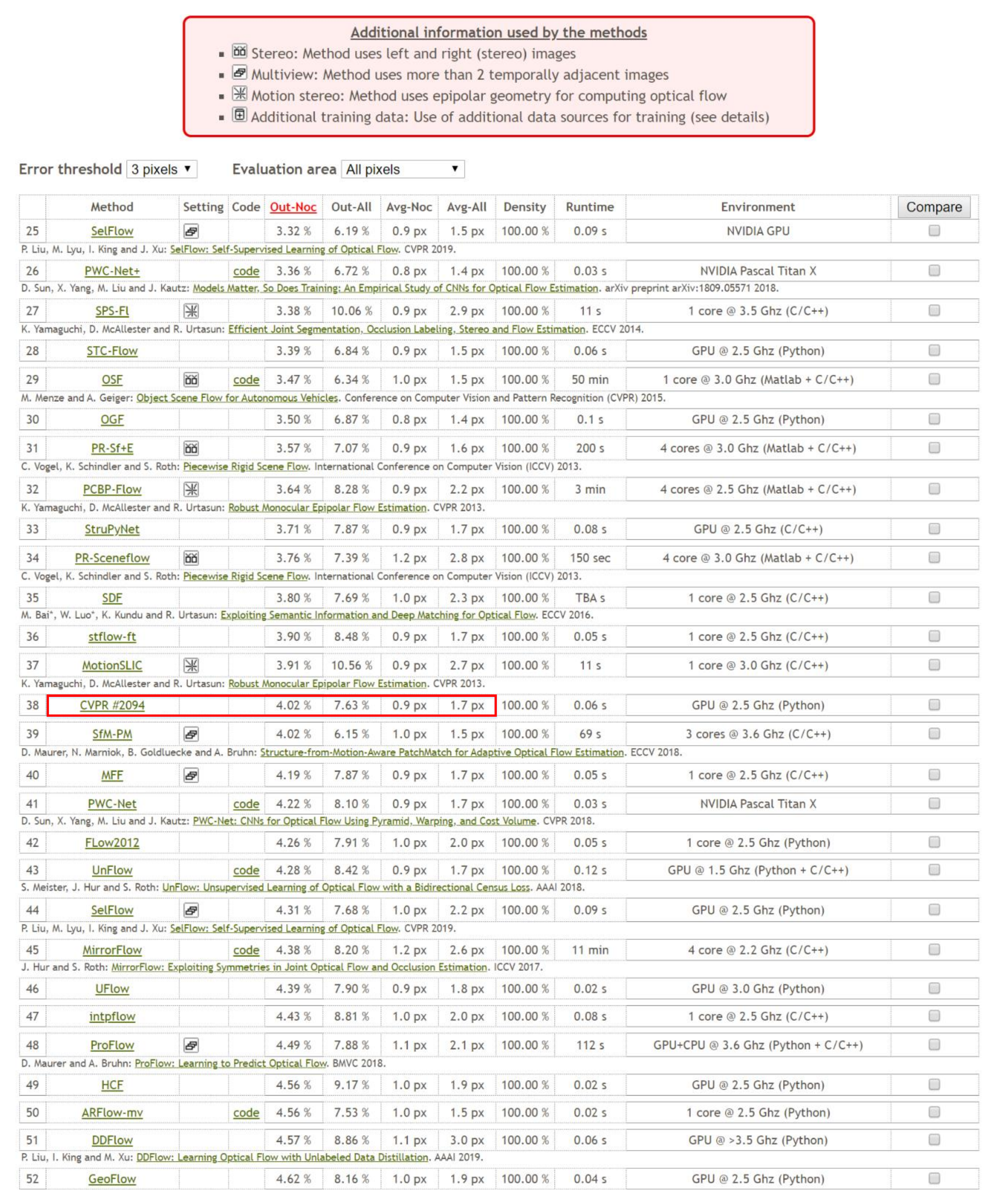}
\caption{Screenshot of KITTI 2012 Optical Flow benchmark on November 15th, 2019. Our unsupervised method achieves comparable performance with supervised methods.}
\label{flow2012}
\end{figure*}

\begin{figure*}[t]
\centering
\includegraphics[width=0.97\textwidth]{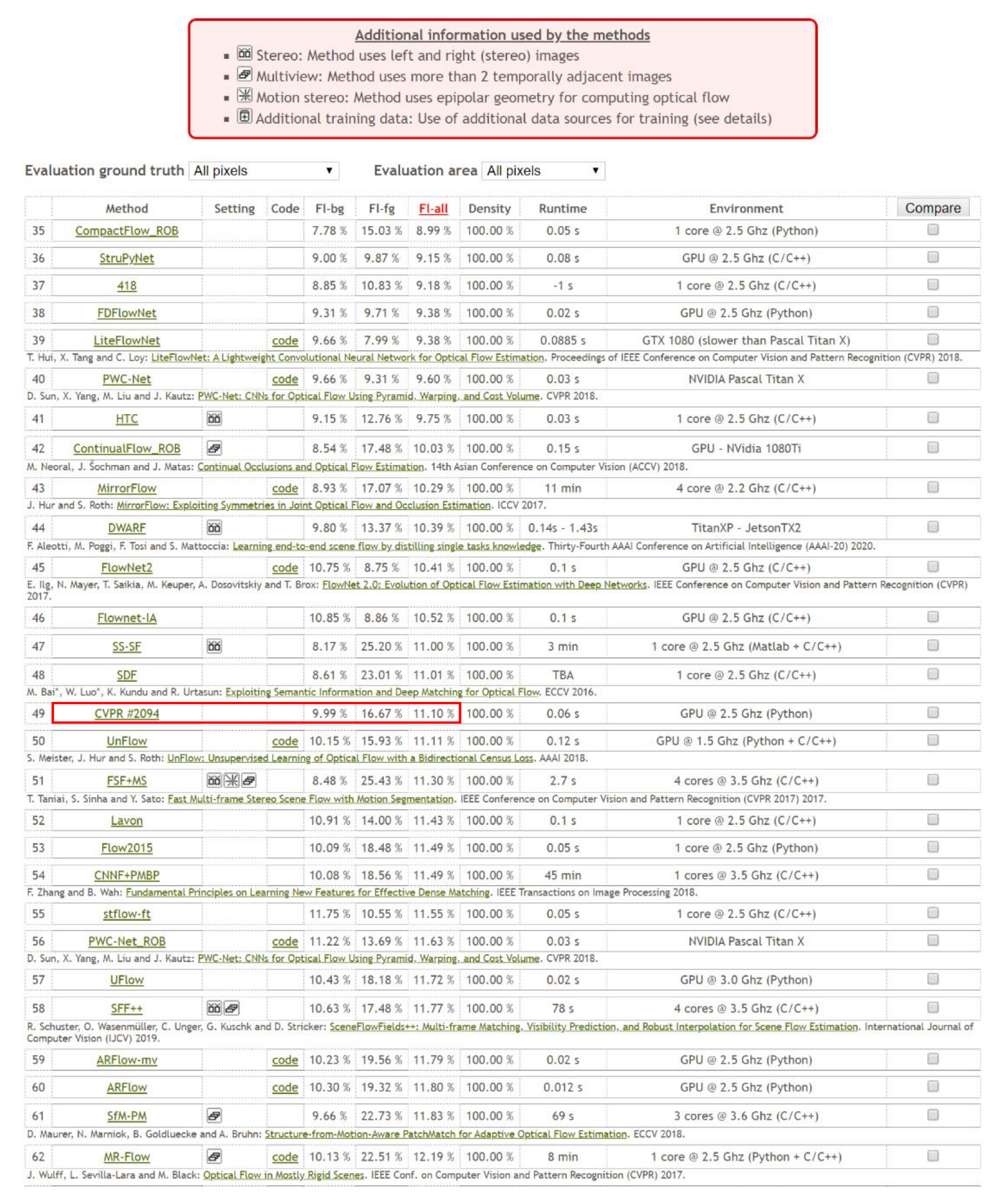}
\caption{Screenshot of KITTI 2015 Optical Flow benchmark on November 15th, 2019. Our unsupervised method achieves comparable performance with supervised methods.}
\label{flow2015}
\end{figure*}

\begin{figure*}[t]
\centering
\includegraphics[width=0.97\textwidth]{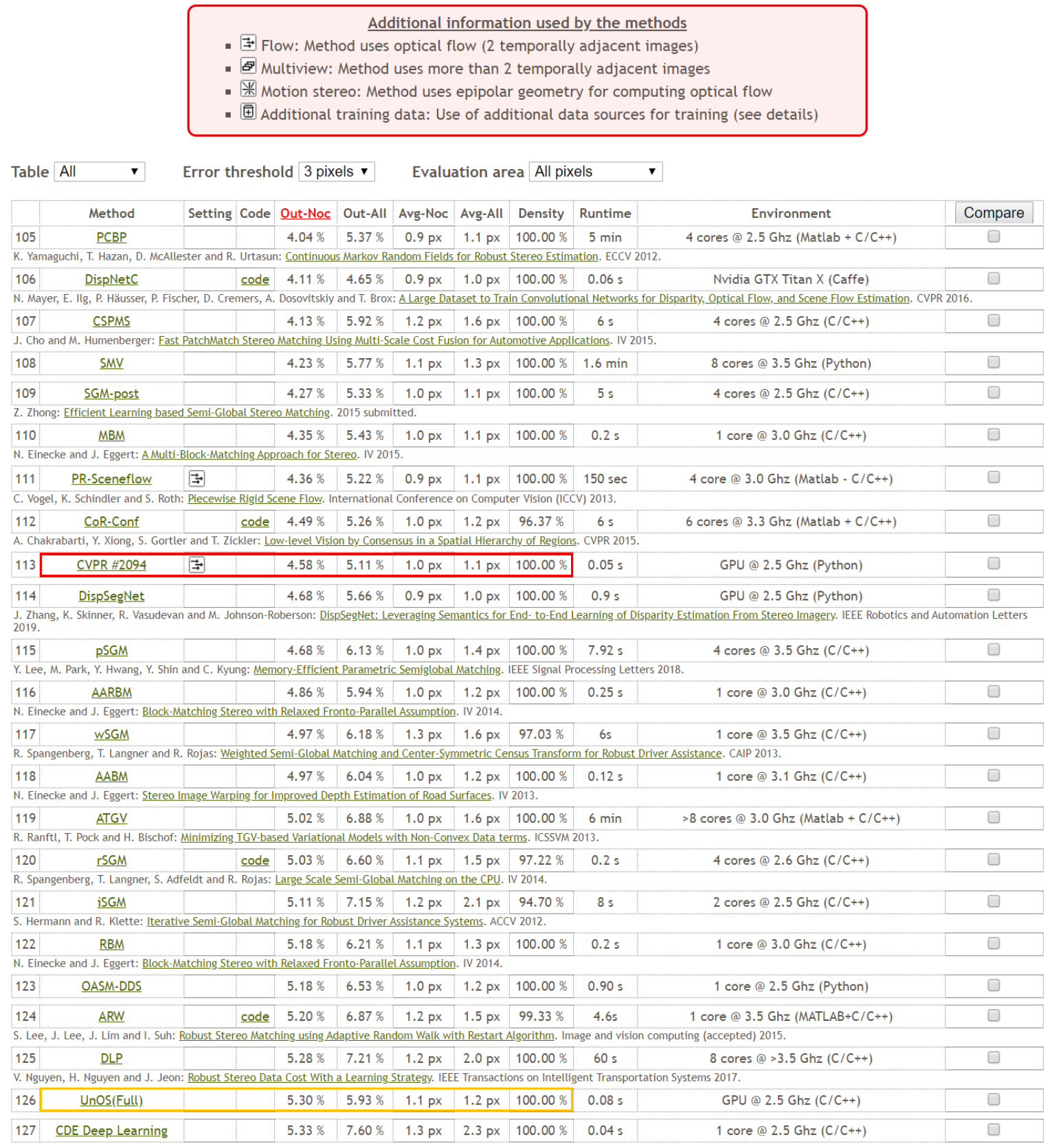}
\caption{Screenshot of KITTI 2012 stereo matching benchmark on November 15th, 2019. We directly estimate stereo disparity with our optical flow model.}
\label{stereo2012}
\end{figure*}

\begin{figure*}[t]
\centering
\includegraphics[width=0.97\textwidth]{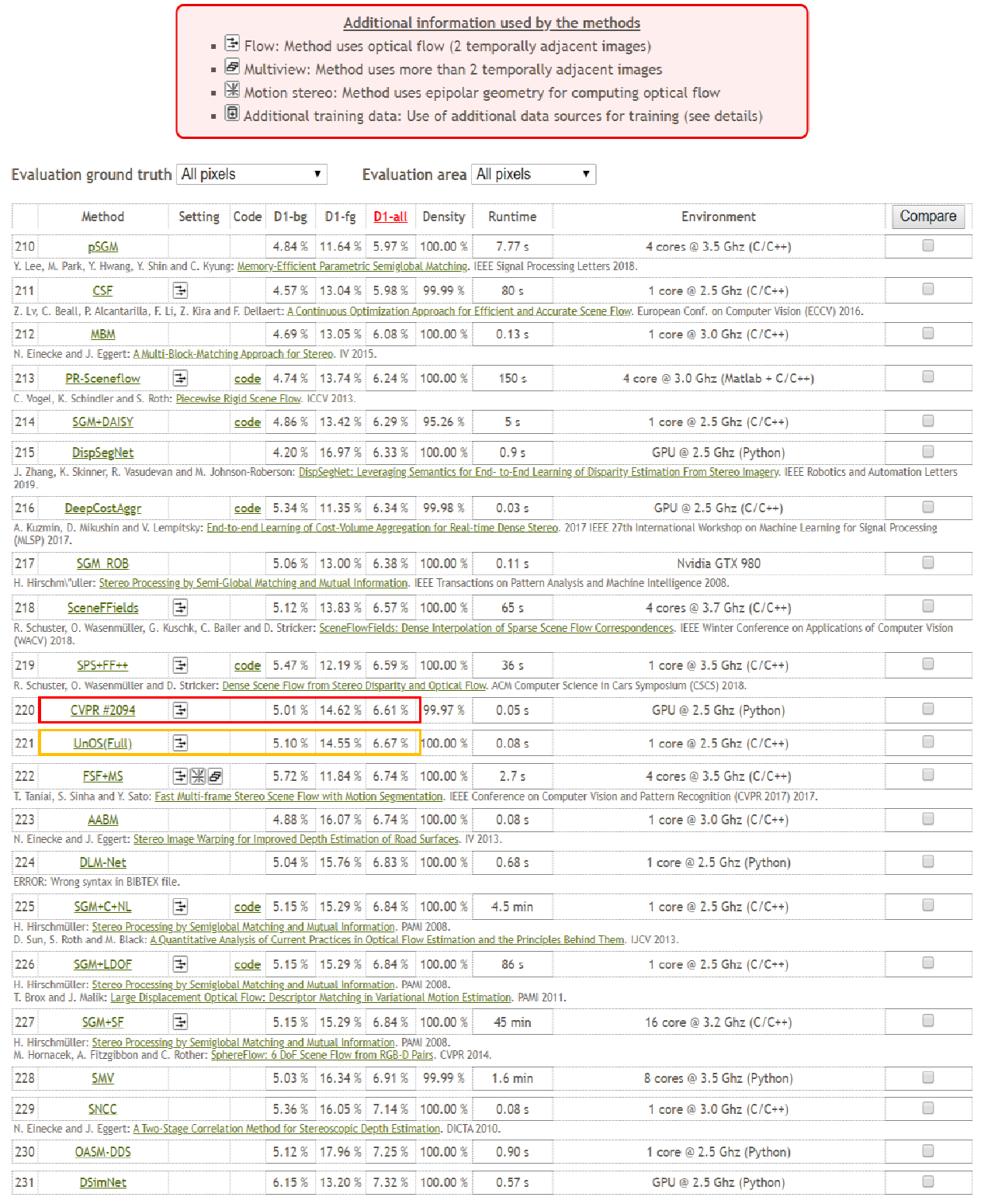}
\caption{Screenshot of KITTI 2015 stereo matching benchmark on November 15th, 2019. We directly estimate stereo disparity with our optical flow model.}
\label{stereo2015}
\end{figure*}

In this supplementary, we show the screenshots on KITTI optical flow and stereo matching benchmarks.
Figure~\ref{flow2012}, Figure~\ref{flow2015}, Figure~\ref{stereo2012}, Figure~\ref{stereo2015} show the screenshots of KITTI 2012 optical flow, KITTI 2015 optical flow, KITTI 2012 stereo matching and KITTI2015 stereo matching benchmarks respectively. Note that on the KITTI benchmarks, some entries on the benchmark utilize additional information (details are shown on the top of the figure). We employ an unified model to predict optical flow and stereo disparity.

For optical flow estimation, our proposed method outperforms all existing unsupervised learning methods and even achieve comparable performance with state-of-the-art supervised learning methods. 

We also submit our stereo matching results to KITTT benchmarks. Our model achieves D1-all = 5.11\% on KITTI 2012 and D1-all = 6.61\% on KITTI 2015, which have similar performance with results on KITTI training sets (table 2 in the main paper). Most entries on KITTI benchmarks are supervised learning methods. The only unsupervised method we can find is UnOS~\cite{wang2019unos}, which is previous state-of-the-art unsupervised stereo matching method. Our method outperforms UnOS~\cite{wang2019unos} on both KITTI 2012 and 2015 stereo matching benchmarks. It deserves to mention that UnOS employs an individual network to estimate stereo disparity, while our method directly employs the optical flow model (unified model) to predict stereo disparity. 	This demonstrates the generalization capability of our model.

\end{document}